\documentclass[conference]{IEEEtran}
\IEEEoverridecommandlockouts

\usepackage{textcomp,amssymb}
\usepackage{setspace}
\usepackage{latexsym,fancyhdr,url}
\usepackage{enumitem}
\usepackage{algpseudocode}
\usepackage{graphics}
\usepackage{xparse} 
\usepackage{xspace}
\usepackage{multirow}
\usepackage{csvsimple}
\usepackage{balance}
\usepackage[vlined,ruled]{algorithm2e}
\usepackage{tikz}
\usepackage[export]{adjustbox}
\usepackage{
  tikz,
  pgfplots,
  pgfplotstable
}
\usepackage{booktabs}
\usepackage{comment}
\usepackage{amsmath}

\DeclareMathOperator*{\argmin}{arg\,min}
\def\BibTeX{{\rm B\kern-.05em{\sc i\kern-.025em b}\kern-.08em
    T\kern-.1667em\lower.7ex\hbox{E}\kern-.125emX}}
\begin{document}

\title{Unlocking Visual Secrets: Inverting Features with Diffusion Priors for Image Reconstruction}
\author{\IEEEauthorblockN{Sai Qian Zhang\IEEEauthorrefmark{1},
Ziyun Li\IEEEauthorrefmark{2},
Chuan Guo\IEEEauthorrefmark{2}, 
Saeed Mahloujifar \IEEEauthorrefmark{2}, 
Deeksha Dangwal\IEEEauthorrefmark{2},\\
Edward Suh\IEEEauthorrefmark{3},
Barbara De Salvo\IEEEauthorrefmark{2} and
Chiao Liu\IEEEauthorrefmark{2}}
\IEEEauthorblockA{\IEEEauthorrefmark{1} New York University, NY, USA, \IEEEauthorrefmark{2} Meta Research, CA, USA, \IEEEauthorrefmark{3} NVidia Research, CA, USA}
}

\maketitle

\begin{abstract}
Inverting visual representations within deep neural networks (DNNs) presents a challenging and important problem in the field of security and privacy for deep learning.
The main goal is to invert the features of an unidentified target image generated by a pre-trained DNN, aiming to reconstruct the original image. Feature inversion holds particular significance in understanding the privacy leakage inherent in contemporary split DNN execution techniques, as well as in various applications based on the extracted DNN features.

In this paper, we explore the use of diffusion models, a promising technique for image synthesis, to enhance feature inversion quality. We also investigate the potential of incorporating alternative forms of prior knowledge, such as textual prompts and cross-frame temporal correlations, to further improve the quality of inverted features. Our findings reveal that diffusion models can effectively leverage hidden information from the DNN features, resulting in superior reconstruction performance compared to previous methods. 
This research offers valuable insights into how diffusion models can enhance privacy and security within applications that are reliant on DNN features.
\end{abstract}


%
\IEEEpeerreviewmaketitle

\section{Introduction}
\label{sec:intro}

Inverting visual features within DNNs presents a significant challenge in the realm of privacy for deep learning. The primary goal of feature inversion is to reverse the outputs (or intermediate results) of a pre-trained DNN and reconstruct the original image. This form of privacy attack, known as~\textit{feature inversion attack}, can raise privacy concerns across various domains. Modern systems that perform face recognition~\cite{azure_face,amazon_face,schroff2015facenet,aggarwal2021fedface,bhat2023face,lezama2017not}, AR/VR applications~\cite{ma2021pixel,fu2023auto,zhang2021f,chu2020expressive}, and image or text retrieval~\cite{zhou2017collaborative,lu2020deep,lu2017hierarchical,song2020information,borgeaud2022improving} often store and process auxiliary data in the form of extracted features from the original input. For example, in a face recognition system, the human face is first encoded with a DNN encoder (e.g., FaceNet~\cite{schroff2015facenet}, CLIP~\cite{bhat2023face,shen2023clip}) and the resultant feature vector is then searched over the database for identity matching via vector comparisons. Feature inversion attacks can be used to reconstruct the face of private users~\cite{mai2018reconstruction}.

\begin{figure}[t]
    \centering
    \includegraphics[width=0.5\textwidth]{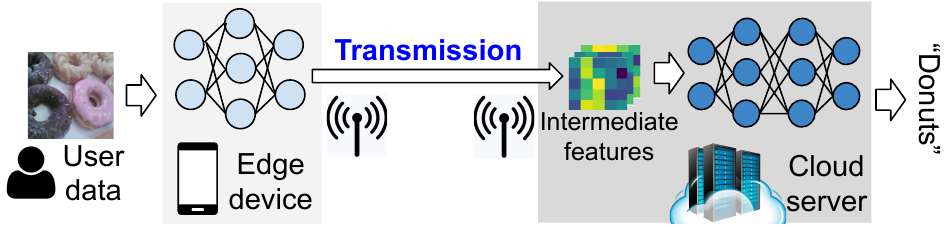}
    \caption{The DNN undergoes layer-wise partitioning and is divided between the edge and the cloud, with the intermediate features being transferred between them. In this example, we focus on the presence of a single edge device.}
    \label{fig:distributed-inference}
\end{figure}

Moreover, feature inversion attack also leads to a serious privacy leakage in the Split DNN computing paradigm~\cite{hauswald2014hybrid, park2022quiltnet, kang2017neurosurgeon,teerapittayanon2016branchynet,teerapittayanon2017distributed,zhang2020adaptive,akintoye2023layer,dong2022splitnets,karjee2022split, luo2023split, mubark2024asap, lee2023wireless, feltin2023dnn, zeng2020coedge, ding2023resource, kang2022dnn, matsubara2019distilled, matsubara2022split, pysyft, splitnn}. Within this paradigm, a layer-wise partitioning of the pretrained DNN into two or more blocks, aligning with the computational capabilities of the edge devices, as shown in Figure~\ref{fig:distributed-inference}. During the execution, the user data is first processed using one or more local DNNs that contain the initial layers. The intermediate results are then transmitted to the central server for the execution of subsequent DNN layers. Split DNN computing has been widely adopted to accommodate the execution of increasingly large DNN on resource-constrained devices like mobile phones, 
and is believed to enhance user privacy by keeping user data on the local device---only the intermediate features are sent to the less secure cloud environment. However, this privacy enhancement turns out to be frail, as recent studies have shown that the intermediate features can be inverted via feature inversion attacks to reconstruct user inputs from the intermediate outputs of parts of the DNN~\cite{mahendran2015understanding,dosovitskiy2016inverting,he2019model,dong2021privacy, maeng2022measuring, song2020information, morris2023text}.

The broad applicability of feature inversion renders it a fundamental problem in ML security and privacy.
On the other hand, feature inversion is not an easy task, particularly when dealing with features extracted from later layers of a network. Intuitively, the learned feature contains more high-level semantic information about objects in an image but less information about the raw input as depth increases. 
As a result, nearly all of the existing feature inversion methods fail when attempting to invert features from later layers of a deep network. This explains why much of the existing research concentrates on feature inversion for shallow DNNs with lower input resolutions~\cite{mahendran2015understanding,dosovitskiy2016inverting,he2019model,maeng2022measuring}.

The recent advancement of generative AI (GenAI) models opens up new possibilities to improve the quality of feature inversion attacks through their comprehensive understanding of image data distributions across real-world scenes.
Among the multitude of existing GenAI techniques, Diffusion Models (DMs)~\cite{ho2020denoising} have emerged as a remarkable breakthrough in generative modeling. Through extensive training with vast datasets comprising millions of real-world images, DMs obtain a high-quality, photorealistic image generation capability. 

In this work, we demonstrate that recent advancements in DMs can be utilized to greatly enhance feature inversion. Instead of inverting DNN features directly to image pixels, we aim to recover the input vector in the latent space of a \emph{latent diffusion model} (LDM) that, when converted to an image through reverse diffusion and forwarded through the DNN, matches the target DNN features. In addition, another noteworthy feature of DMs is their capacity to take textual descriptions as input and produce synthetic outputs conditioned on these textual prompts. We also demonstrate that this capability enables the attackers~\footnote{In this paper, we will employ the terms "attacker" and "adversary" interchangeably.} to specify the prior knowledge of a target image with natural language to utilize their existing knowledge about the victims, if available, by providing a textual description to DMs. Doing so enables the inversion of features that are much deeper into the network.
Finally, in practice, as edge devices often process a continuous stream of input frames,
we propose another variant that uses the temporal correlation in the features between consecutive input frames to enhance the reconstruction quality. Our main contributions are as follows:
\begin{itemize}[leftmargin=*]
\item \emph{Feature inversion using diffusion model prior.} We demonstrate that the exceptional image generation capabilities of DMs can be effectively employed to improve DNN feature inversion. We explore two threat models that closely describe the practical scenarios. To the best of our knowledge, this marks the first research endeavor showing the use of DMs for enhancing DNN feature inversion.
\item \emph{Incorporating textual prior for feature inversion.} We demonstrate that incorporating textual prior information about user inputs can significantly enhance the quality of feature inversion. To integrate this textual prior knowledge and achieve improved feature inversion quality, we introduce new training loss terms as a part of the inversion process.
\item \emph{Feature inversion for videos.} When processing a sequence of temporally correlated inputs, we show that feature inversion can be further enhanced by considering the temporal correlation among consecutive input frames.
\item The evaluation results show that our approach exhibits significant superiority over the state-of-the-art approaches in feature inversion quality across a variety of evaluation metrics. For some backbone DNN models that are trained with self-supervised learning, we can achieve end-to-end inversion by reconstructing the input from the DNN outputs. 
\end{itemize}


\section{Background and Related Works}
\label{sec:background}
\subsection{Diffusion Models}
\label{sec:dm-background}
Diffusion models (DMs)~\cite{ho2020denoising} have recently gained significant attentions for its remarkable ability to generate diverse photorealistic images. It is a parameterized Markov chain trained through Variational inference to generate samples that match the data distribution over a finite duration. Specifically, during the~\textit{forward process} of DMs, given an input image $x_{0}\sim q(x)$, a series of Gaussian noise is generated and added to the $x_{0}$, resulting in a sequence of noisy samples $\{x_{t}\}, 0\leq t \leq T$.
\begin{equation}
    q(x_{t}|x_{t-1}) = \mathcal{N}(x_{t};\sqrt{1-\beta_{t}}x_{t-1}, \beta_{t} I)
\end{equation}
where $\beta_{t}\in (0,1)$ is the variance schedule that controls the strength of the Gaussian noise in each step.

During the~\textit{reverse process}, given a randomly sampled Gaussian noise $\mathcal{N}(x_{T}; 0,I)$, the synthetic images are generated progressively with the following procedure:
\begin{equation}
    p_{\theta}(x_{t-1}|x_{t}) = \mathcal{N}(x_{t-1}; \mu_{\theta}(x_{t},t), \hat{\beta_{t}} I)
\end{equation}
where $\mu_{\theta}(x_{t},t)$ and $\hat{\beta_{t}}$ are defined as follows:
\begin{equation}
    \label{eqn:mean}
    \mu_{\theta}(x_{t},t) = \frac{1}{\sqrt{\alpha_{t}}}(x_{t}-\frac{1-\alpha_{t}}{\sqrt{1-\bar{\alpha}_{t}}}\epsilon_{\theta, t}),\; 
    \hat{\beta_{t}} = \frac{1-\bar{\alpha}_{t-1}}{1-\bar{\alpha}_{t}}
\end{equation}
In equation~\ref{eqn:mean}, $\epsilon_{\theta, t}$ denotes the predicted noise that is generated with a trained U-Net, ${\alpha}_{t}=1-\beta_{t}$, and $\bar{\alpha}_{t}=\prod_{i=1}^{t} {\alpha}_{t}$.

Since their inception, there has been a variety of subsequent research that builds upon DMs. Several alternative approaches to accelerate the reverse process have been proposed~\cite{song2020denoising,lyu2022accelerating,lu2022dpm}. \textit{Latent diffusion models} (LDMs) were introduced in~\cite{rombach2022high} to perform the reverse process within the latent space of an autoencoder. The outcome of this reverse process is then fed to a decoder, which generates the synthetic images. LDMs offer a simple yet efficient way of enhancing both the training and sampling efficiency of LDMs without compromising their quality. LDMs can also be integrated with text encoders for text-to-image generation, as explored in~\cite{saharia2022photorealistic, rombach2022high}. Typically, these models include a pre-trained text encoder that takes user textual descriptions as input, effectively guiding the reverse process to generate the desired synthetic output. 

\subsection{Feature Inversion Attacks against DNNs}
\label{sec:inversion-attacks-background}

\begin{figure}
    \centering
    \includegraphics[width=0.45\textwidth]{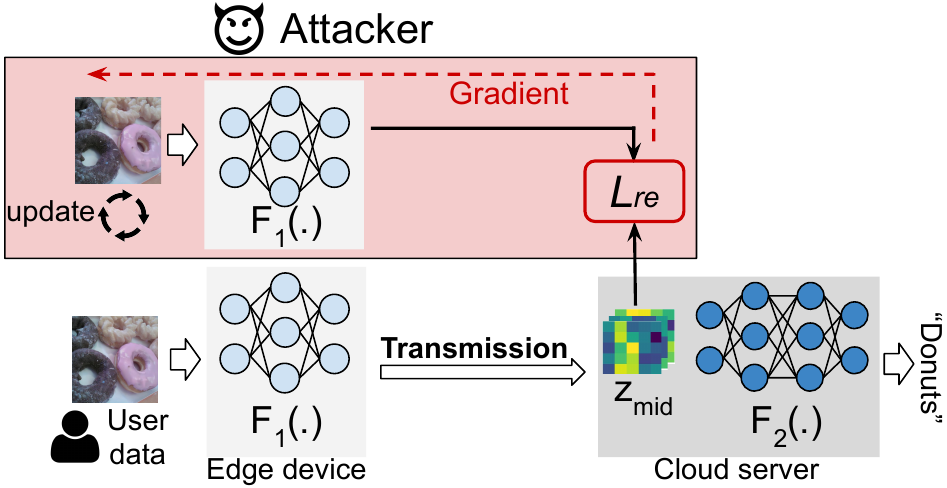}
    \caption{Feature inversion attack against Split DNN execution with white-box settings. $F_{1}(.)$ is exposed to the attacker.}
    \label{fig:feature-inversion}
    \vspace{-0.3cm}
\end{figure}

Feature inversion has been studied by various literature. \cite{dosovitskiy2016inverting} showed that DNN features can be inverted by training a network to reconstruct the corresponding input images given their features. 
\cite{he2019model} first demonstrated that feature inversion can lead to a leakage of private input for split DNN computation, and further showed that introducing the total variation loss~\cite{rudin1992nonlinear} can greatly improve feature inversion quality.~\cite{dong2021privacy} revealed that the exposure of batch normalization parameters can lead to a significant enhancement in feature inversion quality. Unsplit~\cite{erdougan2022unsplit} operated within a black-box setting where the attackers lack knowledge of the model parameters, and developed techniques aimed at reconstructing both user inputs and model parameters.

Our research demonstrates the use of LDMs as a prior can significantly enhance feature inversion quality, an aspect not explored in prior studies. We also explore incorporating diverse prior knowledge sources, such as text and cross-frame correlations, to further improve reconstruction quality. These advanced techniques enable state-of-the-art feature inversion performance, surpassing prior methods. Then, we will discuss various application scenarios of feature inversion attacks.

\subsection{Split DNN Computing}
\label{sec:distributed-inference-background}
Split DNN computing has garnered significant attentions from both academia and industry, as evidenced by numerous studies~\cite{hauswald2014hybrid, teerapittayanon2016branchynet, kang2017neurosurgeon, teerapittayanon2017distributed, karjee2022split, luo2023split, mubark2024asap, lee2023wireless, feltin2023dnn, zeng2020coedge, ding2023resource, kang2022dnn, matsubara2019distilled, matsubara2022split}. Additionally, solutions based on split learning and inference have been actively implemented and embraced across both commercial and open-source applications~\cite{pysyft, splitnn}.
Among the multiple partition strategies~\cite{kang2017neurosurgeon, zhang2020adaptive}, layerwise partition has been widely employed~\cite{hauswald2014hybrid, teerapittayanon2016branchynet, kang2017neurosurgeon, teerapittayanon2017distributed}. This approach entails splitting the DNN into two or more parts and executing on multiple devices.
The study by Hauswald \emph{et al.}~\cite{hauswald2014hybrid}, is among the initial research efforts that moved the later stages of image classification computation to cloud servers. Neurosurgeon~\cite{kang2017neurosurgeon} and DDNN~\cite{teerapittayanon2017distributed} introduced a technique for automatically distributing DNN models between a mobile device and a cloud server, considering factors like network latency and energy usage. Meanwhile, BranchyNet~\cite{teerapittayanon2016branchynet} made use of early exit points within the DNN layers to enable adaptive DNN inference based on the input complexity, further reducing the processing latency.

\subsection{Applications based on Extracted Features}
\label{sec:embedding-loopup}
Modern systems often store and process auxiliary data in the form of features extracted from the DNN encoder. For instance, in face recognition systems, the image of human face is initially encoded with a DNN, and the resulting feature vector is then searched over the database through vector comparisons~\cite{azure_face,amazon_face,schroff2015facenet,aggarwal2021fedface,lezama2017not}.

In addition, some AR/VR tasks, such as Codec Avatar~\cite{ma2021pixel,richard2021audio,fu2023auto}, also rely on extracted features to operate. Codec Avatar is a high fidelity animatable human
face model designed for the purpose of remotely sharing spaces with each other. To generate the Codec Avatar, an encoding process is first performed on the transmitter headset device: cameras linked to the VR headset capture partial facial images, which are then encoded by a DNN model into feature vectors and transmitted to the receiver headset device. On the receiver side, upon the reception of the feature vectors, the decoder reconstructs the avatar's geometry and appearance, enabling the real-time rendering of the transmitter's photorealistic face.

Finally, in the field of image and text retrieval, recent studies have advocated for the adoption of vector database services to facilitate scalable embedding matching and retrieval, yielding enhanced performance~\cite{zhou2017collaborative,lu2020deep,lu2017hierarchical,song2020information,borgeaud2022improving}. To operate, the data owner transmits only embeddings of the raw data from the DNN encoder, to the third-party service, without revealing the actual text content. Subsequently, the database server returns a search result, indicating the index of the matching document on the client side.
\begin{figure}
    \centering
    \includegraphics[width=0.4\textwidth]{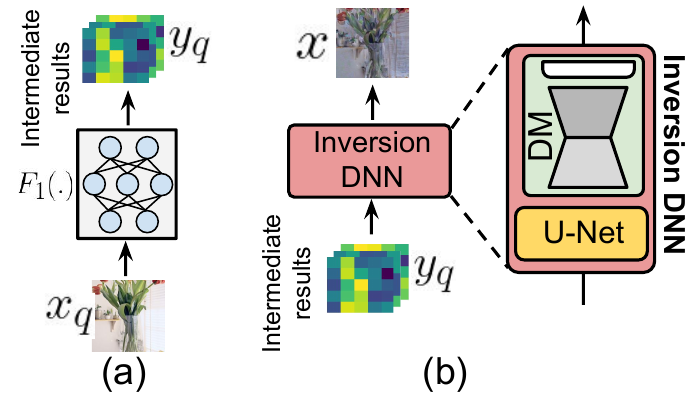}
    \caption{Black-box feature inversion attack procedures.}
    \label{fig:black-box-setting}
\end{figure}


\section{Threat Models}
\label{sec:preliminary}
We begin by first describing the underlying threat model for our feature inversion attack. We consider two settings, white-box and black-box, for the two variants of our attack.

\subsection{Threat Model for White-Box Settings}
\label{sec:white-box-feature-inversion}
We focus on a scenario where the target model $F(.)$ is divided into two parts: $F(.) = F_{2}\circ F_{1}$. Here, we use $x_{gt}$ to represent the user input and $z_{mid}$ to denote the intermediate feature. We make the assumption that $F_{1}$, termed~\textit{user model}, is executed within a secure environment (e.g., edge device) where the intermediate data within $F_{1}$ is protected from any potential leaks to external parties. In contrast, $F_{2}$ operates within an insecure environment, allowing an attacker to access its input $z_{mid} = F_{1}(x_{gt})$. This assumption is reasonable in the context of a split computing scenario such as cloud-edge environment, as the user has control over and can establish trust in the local device's operations. However, in an untrusted cloud environment where either a cloud provider cannot be fully trusted or shared cloud systems have security vulnerabilities, attackers have the potential to exploit vulnerabilities and gain access to the activations for $F_{2}$. Although it is feasible to encrypt the intermediate results, $z_{mid}$, for transmission between the edge and cloud, \textbf{we operate under the assumption that within the insecure public environment (e.g., cloud), the encrypted $z_{mid}$ must be decrypted before further execution}. This introduces vulnerabilities where adversaries could potentially invert $z_{mid}$ and reconstruct the original user input $x_{gt}$. For the white-box setting of the feature inversion attack, we assume the adversary has access to the model structures and parameters of $F_{1}$, and has no prior knowledge of the input $x_{gt}$, nor any intermediate values within $F_{1}$.
Our goal is to reconstruct the input that produces intermediate outcomes resembling $z_{mid}$, which can be formulated as follows:
\begin{equation}
    \label{eqn:white-loss}
    x_{re} = \underset{x}{\argmin} \enspace \mathcal{L}_{re} (F_{1}(x), z_{mid}),
\end{equation}
where $\mathcal{L}_{re}(.,.)$, referred to as the \textit{reconstruction loss}, represents the loss function employed for measuring similarity, with the $l_{2}$ distance being used in this study. We construct $x$ to minimize the loss function as illustrated in Equation~\ref{eqn:white-loss} (Figure~\ref{fig:feature-inversion}). Previous studies~\cite{he2019model,maeng2022measuring} have demonstrated the feasibility of achieving high-fidelity input reconstruction when $F_{1}$ is shallow and the input dimension is limited. However, as $F_{1}$ gets deeper, an increasing portion of information within $x_{gt}$ is filtered out by DNN operations, such as pooling layers, retaining only the essential task-related information. This greatly complicates the task of feature inversion.
\begin{figure}
    \centering
    \includegraphics[width=0.45\textwidth]{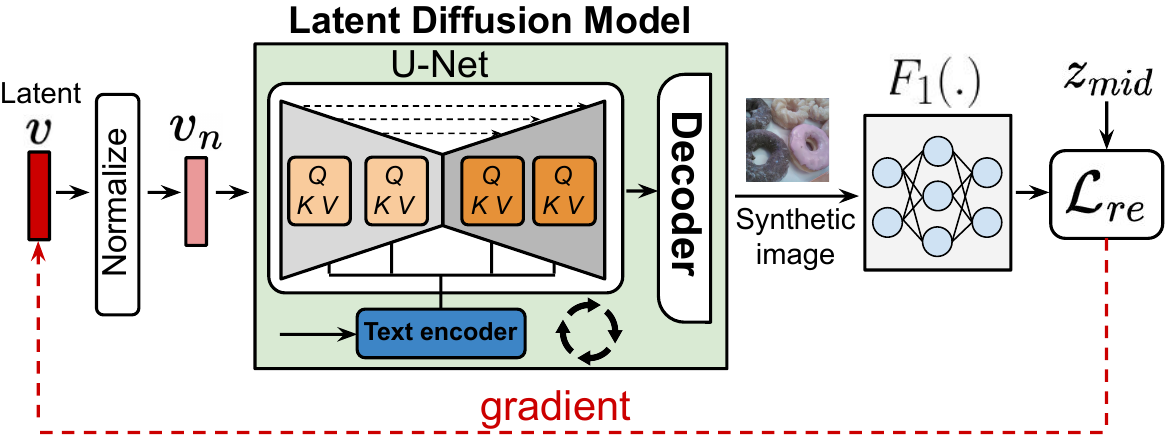}
    \caption{Feature inversion with diffusion model.}
    \label{fig:dm-inversion}
\end{figure}
\subsection{Threat Model for Black-Box Settings}
\label{sec:black-box-feature-inversion}
In the black-box variant of the feature inversion attack, the threat model resembles that of the white-box approach, with the key distinction being the relaxation of the assumption regarding the adversary's knowledge of $F_{1}$(.). Here, the adversary can only gather information about $F_{1}$ indirectly through querying it. As a result, the adversary gains access to the input queries and their corresponding outputs from $F_{1}$(.), as shown in Figure~\ref{fig:black-box-setting}(a). Denote $X_{q} = \{x_{q}\}$ a set of input queries sent by the adversary and $Y_{q} = \{y_{q}\}$ the corresponding outputs from $F_{1}(.)$. 
Next, an inversion DNN $F^{inv}_{\theta}(.)$ is trained to take $Y_{q}$ as input and generate $X$ that closely resembles $X_{q}$ (Figure~\ref{fig:black-box-setting} (b)), namely:  
\begin{equation}
    \label{eqn:black-loss}
    \underset{\theta}{\min} \enspace \sum_{(x_{q}, y_{q})}\mathcal{L}_{re} (F^{inv}_{\theta}(F_{1}(x_{q})), x_{q})
\end{equation}

\subsection{Generalizability of the Threat Model}
\label{sec:applicability_threat_model}
Our threat model described in Section~\ref{sec:white-box-feature-inversion} and Section~\ref{sec:black-box-feature-inversion} can also be applied to systems involving more than two participants. However, given that many real systems are typically divided into two parties~\cite{teerapittayanon2016branchynet,teerapittayanon2017distributed,kang2017neurosurgeon}, we focus on a two-participant system for the remainder of this paper, without sacrificing generality. 

Additionally, by making $F_{1}=F$ and $F_{2}=\varnothing$, our approach can also be applied to the scenario of end-to-end feature inversion, where the objective is to invert the DNN output to reconstruct the input. This presents a significant privacy concern for applications that operate based on extracted features, such as face recognition, Codec Avatar, etc.

\section{White-box Feature Inversion}
\label{sec:white-box-method}

In this section, we describe the white-box attack methodologies in details. 
\begin{algorithm}[t]
\caption{Feature Inversion with LDMs}
\label{alg:feature_inversion_alg}
\DontPrintSemicolon
  \KwIn{
  $F_{1}(.)$ is the user DNN model. \\ 
  $v$ is the input latent vector of LDMs. \\  
  $I$ is total number of iterations. \\
  $\epsilon$ is the learning rate. \\ 
  } 
  \small
  \For{$1 \leq i \leq I$}{  
       {$v^{i}_{n} = \frac{v^{i}-\mathrm{mean}(v^{I})}{\mathrm{std}(v^{i})}$} \\
       {$\mathcal{L}_{tot} = ||F_{1}(D(v^{i}_{n}))-z_{mid}||^{2}$ + $\lambda_{s}TV(D(v^{i}_{n}))$} \\
       {$v^{i+1} = v^{i} - \epsilon \frac{d\mathcal{L}_{tot}}{dv}$} \\
       {$i = i + 1$} \\
   }
    {$v_{n} = \frac{v^{I}-\mathrm{mean}(v^{I})}{\mathrm{std}(v^{I})}$} \\
   {\Return $D(v_{n})$}.
\end{algorithm}
We leverage the prior knowledge embedded within the LDM to reconstruct the user input $x_{gt}$. Let $D(v, e)$ represent the generating function of the LDM. Here, $v$ denotes the input latent variable, which is expected to follow a normal distribution~\cite{ho2020denoising}, and $e = E(t)$ represents the text embedding. The function $E(.)$ indicates a pre-trained text encoder, and $t$ corresponds to the text input provided by the user. In this section, we ignore the text prompt by setting the text embedding $e$ to a vector of zeroes, and will examine the influence of the text prior in Section~\ref{sec:White-box-text-prior}. We then search for the input latent variable $v$ that allows the LDM to produce a synthetic output, denoted as $D(v_n)$. This output, when passed to $F_{1}$(.), will result in a similar intermediate output as $z_{mid}$ (Figure~\ref{fig:dm-inversion}).
\begin{equation}
    \label{eqn:dm-loss}
    v^{*} = \underset{v}{\argmin} \enspace \mathcal{L}_{re}(F_{1}(D(v_{n})), z_{mid}) + \lambda_{s}TV(D(v_{n}))
\end{equation}
As the LDM necessitates input data to approximate a normal distribution for photorealistic image generation, we implement a soft restriction on the variable $v$ by normalizing it prior to forwarding it to the LDM. Specifically, we define $v_{n} = \frac{v-\mathrm{mean}(v)}{\mathrm{std}(v)}$ as the normalized version of $v$, which serves as the input for the LDM. We observe that applying normalizing operation can greatly enhance the feature inversion performance.
$z_{mid} = F_{1}(x_{gt})$ is the intermediate result generated from the user input. $TV(.)$ represents the~\textit{Total Variation}~\cite{rudin1992nonlinear} which is used to reduce the abrupt pixel variations across the reconstructed image. $TV(x)$ is defined as follows:
\begin{equation}
    \label{eqn:tv}
    TV(x) = \frac{1}{MN}\sum_{i}\sum_{j}(|x_{i+1,j}-x_{i,j}|^{2} + |x_{i,j+1}-x_{i,j}|^{2})
\end{equation}
where $M$ and $N$ represent the spatial size of the image, and $\lambda_{s}$ denote the weight of the TV loss. The feature inversion process is summarized in Algorithm~\ref{alg:feature_inversion_alg}.

\subsection{White-box Inversion with Textual Prior}
\label{sec:White-box-text-prior}

Another important characteristic of LDMs is their ability to take text prompt as input and produce synthetic outputs guided by textual descriptions. We leverage this capability in our feature inversion attacks by allowing the attacker to express their prior knowledge about the user input in the form of natural language. Different from generic image priors such as total variation, this form of text prior can be specific to each target image and further enhances the quality of feature inversion using diffusion models.

\begin{figure}
    \centering
    \includegraphics[width=0.4\textwidth]{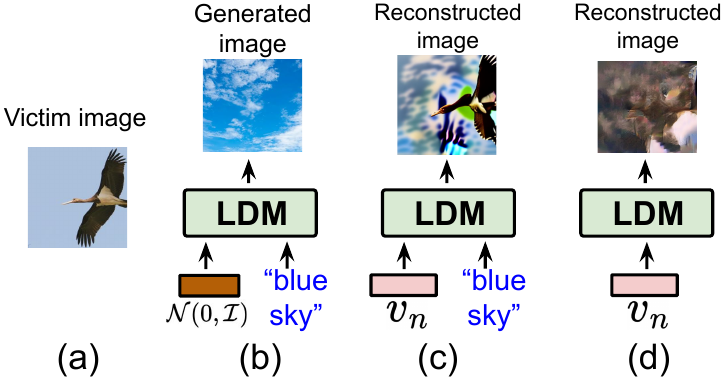}
    \caption{Impact of textual prior on feature inversion.}
    \label{fig:textual_prior_illustration}
\end{figure}

To incorporate the text prior into the feature inversion process, consider the private user image depicted in Figure~\ref{fig:textual_prior_illustration} (a). Assuming the adversary possesses prior knowledge of this private image, they will convey this knowledge to the LDM through textual description. The LDM will take these textual description together with another randomly generated, normally-distributed input to produce an image visually akin to the user image (Figure~\ref{fig:textual_prior_illustration} (b)). Subsequently, we proceed to further train the LDM input $v_{n}$ to enhance the LDM's ability to refine the output, making it more closely resemble the user image $x_{gt}$ (Figure~\ref{fig:textual_prior_illustration} (c)). The resulting reconstructed image has a much better quality than that reconstructed without a textual description, which is shown in Figure~\ref{fig:textual_prior_illustration} (d).

\begin{figure}
    \centering
    \includegraphics[width=0.47\textwidth]{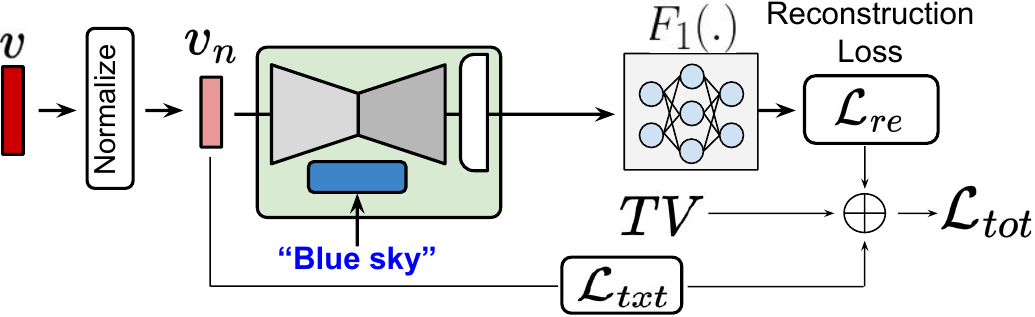}
    \caption{White-box feature inversion with textual prior.}
    \label{fig:textual_prior}
\end{figure}

Considering that the normally-distributed latent coupled with the text prior can produce an output that relates to user input, we utilize this insight by further pushing $v_n$, the input of LDM, to approach a random variable generated from a normal distribution. To achieve this, we assess the Gaussianity of $v_{n}$ using the negentropy metric outlined in~\cite{hyvarinen2000independent}, resulting in an additional loss term denoted as $\mathcal{L}_{txt}$, defined as follows:
\begin{equation}
\label{eqn:text-prior-loss-only}
    \mathcal{L}_{txt} = -\mathop{\mathbb{E}} \Bigl[\frac{1}{\alpha^{2}}log cosh^{2}(\alpha v_{n,i}) \Bigr]
\end{equation}
where $1\leq \alpha \leq 2$ is a hyperparameter, the expectation $\mathop{\mathbb{E}}(.)$ is taken over the elements of $v_{n}$. The total loss $\mathcal{L}_{tot}$ can be described as:
\begin{equation}
\label{eqn:text-prior-loss}
    \mathcal{L}_{tot} = \mathcal{L}_{re}(F_{1}(D(v_{n}, e)), z_{mid}) + \lambda_{s}TV + \lambda_{txt} \mathcal{L}_{txt}(v_{n})
\end{equation}
where $e=E(t)$ represents the embedding of the textual description, which serves as an additional input to the LDM, $\lambda_{txt}$ is the weight factor to balance the loss terms (Figure~\ref{fig:textual_prior}). The remaining inversion algorithm is similar to Algorithm~\ref{alg:feature_inversion_alg}. The detailed algorithm is given in the appendix.

\subsection{White-box Multi-frame Reconstruction}
\label{sec:white-box-multi-frame}

In this section, we explore the problem on multi-frame feature inversion. This scenario closely resembles real-world situations where edge devices handle a continuous stream of input frames, such as burst mode photos or video clips. In this context, the local DNN processes consecutive input frames that exhibit temporal correlation, the intermediate features are then transmitted to cloud servers for subsequent processing. The goal is to reconstruct the entire input image sequence using the intermediate results.

In particular, consider $x_{gt,k}$, where $1\leq k\leq K$, to represent a sequence of $K$ user inputs. Additionally, let $z_{mid,k}$ and $v_{k}$ represent the corresponding local DNN output and input latent variable for $x_{gt,k}$. We introduce an additional loss component $\mathcal{L}_{c}(.)$ aimed at minimizing the disparity among the latent vectors $v_{k}$ across these frames. To achieve this, the multi-frame reconstruction process can be realized by solving the following optimization problem:

\begin{figure*}
    \centering
    \includegraphics[width=0.85\textwidth]{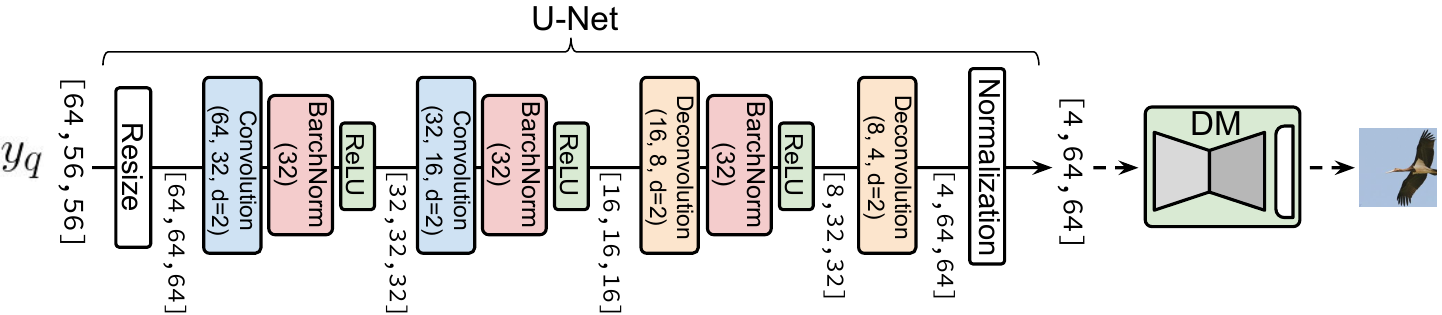}
    \caption{A sample architecture of U-Net. The dimension will adjust based on the dimension of $y_{q}$. In this example, the intermediate result $y_{q}$ is of dimension $64\times 56\times 56$, with a batch size of 1. The LDM input has a dimension of $4\times 64\times 64$. $d$ in the (de)convolutional blocks denotes the stride.}
    \label{fig:feature-inversion-arch}
\end{figure*}

\begin{equation}
    \label{eqn:temporal-loss}
    \underset{v_{k, 1\leq k\leq K}}{\min} \sum_{k = 1}^{K} \Bigl[\mathcal{L}_{re,k} + \lambda_{s}TV_{k} + \lambda_{c} \mathcal{L}_{c}(v_{k}, \bar{v})\Bigr],
\end{equation}
where $\bar{v} = \frac{1}{K}(\sum_{k = 1}^{K} v_{k})$ represents the average of the input latent vectors across the $K$ frames. The loss function $\mathcal{L}_{c}$ is utilized to minimize the disparity between the latent vectors for each frame. In this study, we have observed that simply minimizing their $l_{2}$ distance yields an excellent reconstruction quality. The parameter $\lambda_{c}$ serves as a weight to balance the importance of these two loss functions. $\mathcal{L}_{re,k}$ and $TV_{k}$ denote the reconstruction loss and total variation loss for reconstructing $x_{gt,k}$. The detailed algorithm is given in the appendix.

\section{Black-box Feature Inversion}
\label{sec:black-box-method}
\begin{figure}
    \centering
    \includegraphics[width=0.4\textwidth]{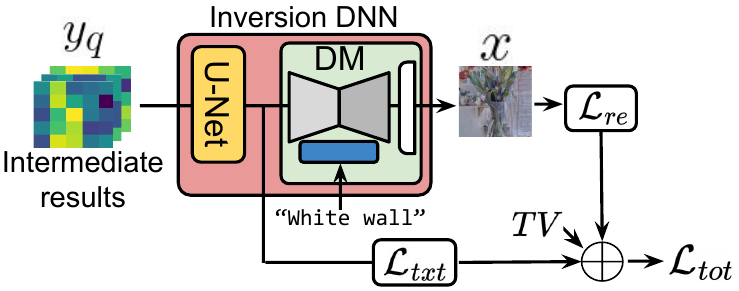}
    \caption{Feature inversion with textual prior under black-box settings.}
    \label{fig:feature-inversion-txt-prior}
\end{figure}
For feature inversion attacks with black-box settings, the attacker obtains access to the input queries and their corresponding outputs from $F_{1}$(.), as depicted in Figure~\ref{fig:black-box-setting} (a). Let $X_{q} = \{x_{q}\}$ denote a set of input queries sent by the adversary and $Y_{q} = \{y_{q}\}$ represent the corresponding outputs from $F_{1}(.)$. The adversary then proceeds to train an inversion DNN $F^{inv}_{\theta}(.)$ designed to take $y_{q}$ as input and generate $x$ that closely resembles $x_{q}$ (Figure~\ref{fig:black-box-setting} (b)).

$F^{inv}_{\theta}(.)$ consists of two major components: a pre-trained LDM and an U-Net, which are denoted as $D(.)$ and $F_{u}(.)$, respectively. During the execution, $F_{u}(.)$ takes the intermediate data $y_{q}$ and generates the input latent variable for the LDM, which then produces the result $x = D(F_{u}(y_{q}))$. The training of the inversion DNN model involves minimizing the following loss function:
\begin{equation}
    \label{eqn:dm-black-loss}
    \theta_{u}^{*} = \underset{\theta_{u}}{\argmin} \enspace \sum_{(x_{q},y_{q})\in \{(X_{q}, Y_{q})\}}\mathcal{L}_{re}(D(F_{u}(y_{q})), x_{q}) + \lambda_{s} TV
\end{equation}
where $\theta_{u}$ represents the parameters of $F^{inv}_{\theta}(.)$. TV loss is introduced over the reconstructed input $D(F_{u}(y_{q}))$.

The architecture of U-Net is illustrated in Figure~\ref{fig:feature-inversion-arch}. During the forward pass, the input $y_{q}$ is first resized spatially. Following this, the intermediate output traverses through several blocks consisting of (de)convolutional layers, batch normalization layers, and ReLU layers. The resulting output is normalized before being sent to the diffusion model for image generation. The specific dimensions of the inversion DNN will vary depending on the shape of $y_{q}$.

\subsection{Black-box Inversion with Textual Prior}
\label{sec:black-box-textual-prior}

Similar to the incorporation of textual priors to enhance reconstruction quality in the white-box setting, integrating textual priors into the inversion DNN can also improve the quality of the black-box feature inversion. The training procedure is outlined in Figure~\ref{fig:feature-inversion-txt-prior}. The U-Net output is directed into the DM for image generation, while an additional loss function $\mathcal{L}_{txt}(.)$ is simultaneously applied to enhance its gaussianity. The overall loss function is shown as follows:
\begin{equation}
    \scriptsize
    \label{eqn:dm-black-loss-text-prior}
    \theta_{u}^{*} = \underset{\theta_{u}}{\argmin} \enspace \sum_{(x_{q},y_{q})}\mathcal{L}_{re}(D(F_{u}(y_{q}), e_{q}), x_{q}) + \lambda_{s} TV + \lambda_{txt} \mathcal{L}_{txt}(F_{u}(y_{q}))
\end{equation}
where $\mathcal{L}_{txt}(.)$ is the loss term that enforces the LDM inputs, $F_{u}(y_{q})$, follows a gaussian distribution. $e_{q}=E(t_q)$ is the embeddings of the textual description $t_{q}$ that describes $x_{q}$.

\subsection{Black-box Multi-frame Reconstruction}
\label{sec:black-box-multi-frame}

In this section, we explore the problem on multi-frame feature inversion under black-box settings. In particular, assume a group of consecutive frames with a total of $K$ images, the inversion DNN will take the intermediate results $Y^{g}_q = \{y_{q,k\in K}\}$ from each frame $k$ within this group $g$, and produce the $K$ outputs that will serve as the inputs of the LDM.

In order to exploit the temporal correlation within the intermediate results $Y^{g}_q$, we introduce a pointwise convolutional layer into the inversion DNN, as depicted in Figure~\ref{fig:inversion-dnn-multiframe}. This pointwise convolutional layer will incorporate a weight filter with a spatial size of $1\times 1$, enabling the learning of temporal correlations among the intermediate results $y_{q,k}$. The output of the pointwise convolution will then be separated into $K$ pieces each of which corresponds to one input frame, the outputs will be forwarded to the U-Net, whose architecture is shown in Figure~\ref{fig:feature-inversion-arch}. Each of the four U-Net will share its weights. The outputs from U-Net will further be delivered to the LDM, which will then reconstruct $x_{q,k}$ for each $k\in K$. The loss function for training is shown as follows:
\begin{figure}
    \centering
    \includegraphics[width=0.4\textwidth]{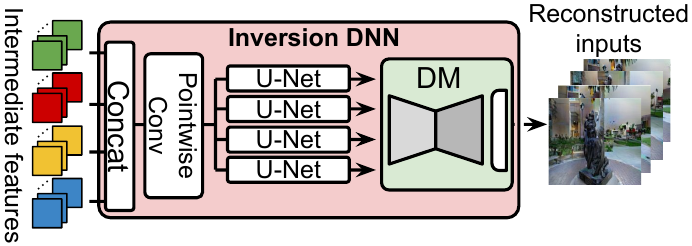}
    \caption{Architecture of the inversion DNN for black-box multi-frame reconstruction. In this example, the inversion DNN can generate four frames.}
    \label{fig:inversion-dnn-multiframe}
\end{figure}
\begin{equation}
    \label{eqn:black-multi-frame-loss}
    \underset{\theta_{u}}{\min}\enspace \sum_{g}\big[\mathcal{L}_{re}(D(F_{u}(Y^{g}_{q})), X^{g}_{q}) + \sum_{k\in K}\lambda_{s} TV^{g}_{k}\big]
\end{equation}

where $X^{g}_{q} = \{x_{q,k\in K}\}$ are the groups of ground-truth consecutive frames, and $TV_{k}$ is the TV loss of the $k$-$th$ reconstructed frame within group $g$.

\section{Evaluation Results for White-box Inversion}
\label{sec:eval}
In this section, we present detailed evaluation of the white-box feature inversion technique described in Section~\ref{sec:white-box-method}. We first evaluate the quality of the inverted features over different applications in Section~\ref{sec:inversion-diffusion-eval}, Section~\ref{sec:white-box-eval-object-detection} and Section~\ref{sec:white-box-clip-inverseion}. Next, we explore the influence of the textual context in Section~\ref{sec:text-prior-eval} and the multi-frame reconstruction in Section~\ref{sec:multi-frame-results}. Lastly, we conduct an ablation study on the number of diffusion sampling steps in Section~\ref{sec:ablation}.

\subsection{Experiment settings}
\label{sec:white-box-settings}
\textbf{Datasets and models:} We assess our feature inversion approach outlined in Algorithm~\ref{alg:feature_inversion_alg} on ImageNet~\cite{deng2009imagenet} and COCO~\cite{lin2014microsoft} datasets. We employ various DNN architectures pre-trained on ImageNet as the target models for feature inversion, including ResNet-18, ResNet-50, and Vision Transformer (ViT). All of the pretrained models are downloaded from the official Pytorch website. Due to institutional restriction, we use a LDM whose architecture is similar to Stable Diffusion 2.1~\cite{rombach2022high} and is trained on a proprietary licensed datasets of image-caption pairs. Note that our approach can also be applied to other LDMs.

\textbf{Hyperparameters:} We set all $\lambda_{s}$ to 1 for the reconstruction loss defined in equation~\ref{eqn:dm-loss}, equation~\ref{eqn:text-prior-loss}, and equation~\ref{eqn:black-multi-frame-loss}. The reconstruction process continues for a total of $T=1500$ iterations. We adopt the Adam optimizer with an initial learning rate of 0.1, $\beta$=(0.9,0.999). To expedite the reverse procedure, we configure the sampling steps of LDM to be 20 with a linear schedule~\cite{ho2020denoising}. We find that using 20 sampling steps can already yield high-quality feature inversion results. We investigate the impact of the sampling steps in Section~\ref{sec:ablation}. More evaluation results can be found in the appendix.
\begin{figure}
    \centering
    \includegraphics[width=0.4\textwidth]{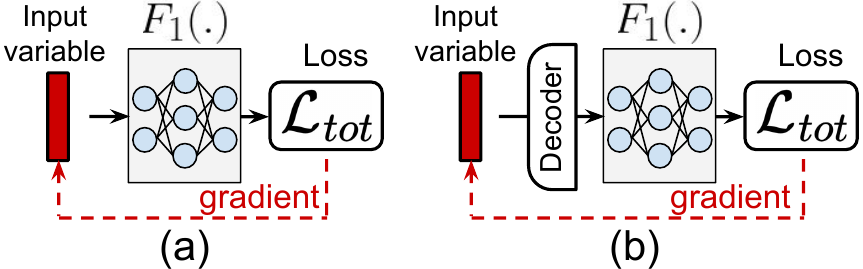}
    \caption{Baseline feature inversion algorithms.}
    \label{fig:baselines}
\end{figure}

\textbf{Baselines:} We consider two algorithms for comparison. The first approach, referred to as ~\textit{Direct Optimization (DO)}, reconstructs the input by directly optimizing Equation~\ref{eqn:white-loss} over the image pixel space (Figure~\ref{fig:baselines} (a)). This method has been utilized for input reconstruction in prior works~\cite{he2019model,maeng2022measuring, erdougan2022unsplit} and serves as the baseline to assess the impact of LDMs on the feature inversion. The second approach, known as the~\textit{Decoder-based (DB)} approach (Figure~\ref{fig:baselines} (b)), employs only the LDM decoder for input reconstruction. Evaluating this approach helps us understand the influence of the iterative reverse process in LDMs on feature inversion attacks. It is worth noting that this \textbf{Decoder-based approach has not been investigated in prior works on feature inversion either}, but similar techniques using GAN decoders have been used in the context of gradient inversion~\cite{jeon2021gradient, Li_2022_CVPR} and model inversion~\cite{Zhang_2020_CVPR}. Finally, we denote our method as the~\textit{DM-based (DMB)} approach.

\subsection{Feature Inversion on Split Models for Image Classification}
\label{sec:inversion-diffusion-eval}
We first assess the reconstruction quality of our feature inversion attacks \emph{without} text prior. We randomly select 100 images from the ImageNet and COCO test datasets and feed them to a pre-trained ResNet-18, ResNet-50 and ViT-base model, respectively. For each of these target DNN models, we evenly divide them into blocks of layers and extract intermediate results at the end of each block. Subsequently, we employ the techniques outlined in Section~\ref{sec:white-box-method} to reconstruct the user input.

\begin{figure*}
    \centering
    \includegraphics[width=0.8\textwidth]{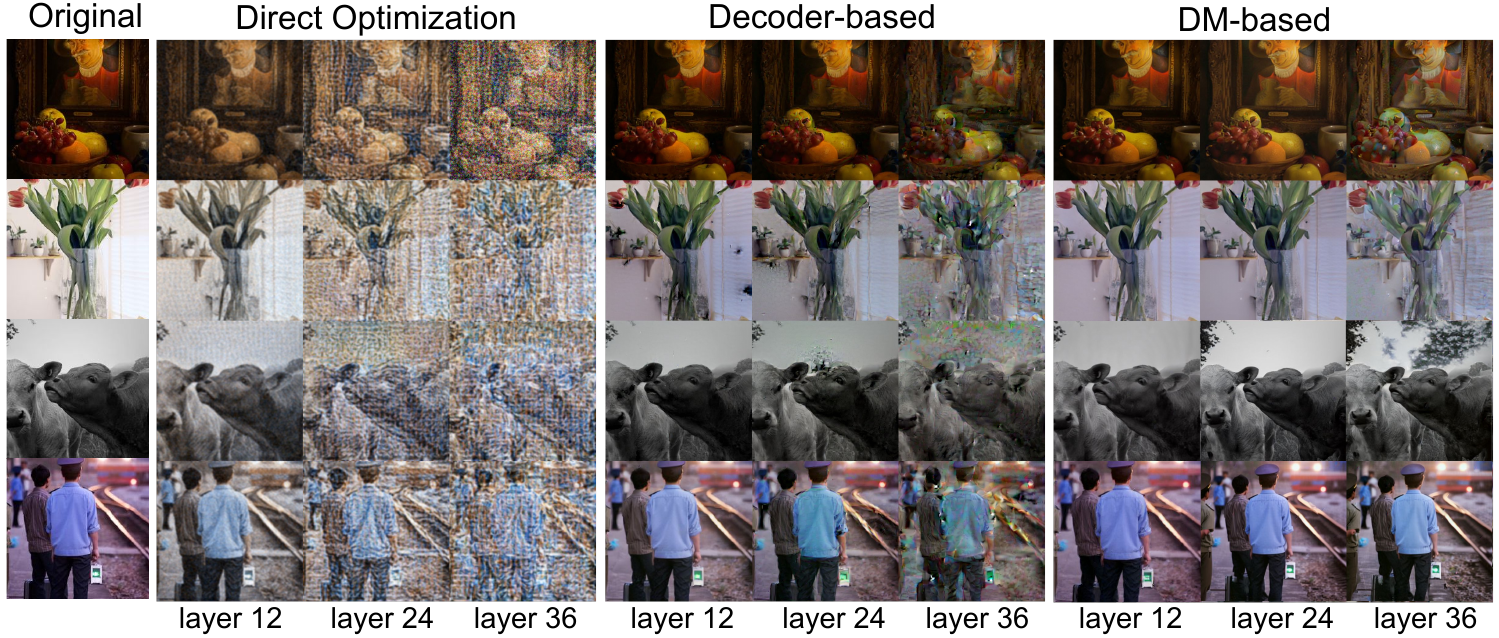}
    \caption{Feature inversion over ResNet-50 on ImageNet, features are extracted from the end of layer 12, layer 24 and layer 36 respectively. }
    \label{fig:resnet50-coco-in-paper}
\end{figure*}

\textbf{Qualitative result:} Figure~\ref{fig:resnet50-coco-in-paper} depicts the feature inversion results for ResNet-50 over ImageNet, respectively. The original image is displayed in the left column for reference. DM-based method consistently demonstrates superior reconstruction qualities across all datasets and DNN architectures. Notably, our approach achieves high-quality input reconstructions, even when utilizing features from very deep layers (e.g., layer 36 in ResNet-50), whereas other baseline methods struggle to achieve comparable performance. 

\textbf{Quantitative result:} To quantify the quality of the reconstructed images, we utilize three metrics. The first metric is Inception Score (IS)~\cite{salimans2016improved}, which is commonly used to evaluate the quality of image generation in prior works~\cite{song2020analyzing,xu2019ganobfuscator,he2019model,dong2021privacy}.  For instance, generative AI models like StackGAN~\cite{zhang2017stackgan} and GAN-INT-CLS~\cite{reed2016generative} typically generate images with IS scores around 3 to 5, while diffusion models can achieve IS scores as high as 10~\cite{ho2020denoising}. 

The second metric is Peak Signal-to-Noise Ratio (PSNR), which calculates the ratio between the maximum possible value of a signal and the power of distorting noise affecting the quality of its representation. The mathematical expression for PSNR is as follows:
\begin{equation}
\label{eqn:psnr}
PSNR(I^{ori}, I^{re}) = 10log(\frac{255}{\frac{1}{MN}\sum_{i=1}^{M}\sum_{j=1}^{N}I^{ori}_{i,j}-I^{re}_{i,j}})
\end{equation}
where $I^{ori}$ and $I^{re}$ denote the original and reconstructed images, and they both have a size of $M\times N$. $255$ is the maximum pixel value. PSNR is a commonly used metric for assessing image quality, particularly when comparing a compressed or reconstructed image to its original version. It is frequently employed in the image processing tasks to quantify the degree of distortion introduced. PSNR values between 30-50 dB are typically considered indicative of excellent image quality.


Lastly, the Structural Similarity Index Measure (SSIM)~\cite{wang2004image} is a metric used to assess image quality by comparing the similarity between two images. In our scenario, we evaluate SSIM between the reconstructed images and the original image. Unlike the PSNR, which focuses solely on pixel differences, SSIM evaluates changes in structural information, luminance, and contrast, making it more closely aligned with human visual perception. SSIM values range from 0 to 1, with higher values indicating better image quality.

\begin{table}
 \centering
 \caption{Evaluation results: 'DO', 'DB', and 'DMB' refer to direct optimization, decoder-based, and DM-based approaches. 'L' is the feature extraction layer. PSNR is shown in db. For IS, PSNR and SSIM, higher values indicate better results.}
\begin{adjustbox}{width=\columnwidth,center}
    \begin{tabular}{lcccccccccccc}\toprule 
    & \multirow{2}{*}{\begin{tabular}{@{}c@{}}Metric\end{tabular} } & \multirow{2}{*}{\begin{tabular}[c]{@{}c@{}}Method\end{tabular} }
    & \multicolumn{3}{c}{ResNet-18} & \multicolumn{3}{c}{ResNet-50} & \multicolumn{3}{c}{ViT-base}  \\
    \cmidrule(lr){4-6} \cmidrule(lr){7-9} \cmidrule(lr){10-12}
    &  &  & L4  & L8  & L12  & L12  & L24  & L36  & L3  & L4  & L5 \\ \toprule
    & & DO   & 5.63 & 3.92  & 1.40 & 5.55  & 3.88 & 1.28 & 5.46 & 3.95 & 1.63 \\
    &IS & DB  & 6.84 & 5.97  & 4.20 & 6.90  & 5.86 & 4.38 & 6.76 & 5.76 & 3.93 \\
    & & DMB   & 7.23 & 6.86  & 6.48 & 7.36  & 6.90 & 6.55 & 7.14 & 6.77 & 6.58 \\\midrule
    & & DO   & 29.3 & 14.6  & 9.51 & 28.9   & 15.3 & 8.04 & 28.5  & 17.6 & 9.42 \\
    &PSNR & DB  & 35.2 & 32.6  & 18.6 & 35.4   & 33.0 & 18.3 & 36.8   & 33.1 & 19.9 \\
    & & DMB   & 41.0 & 36.3  & 29.1 & 40.2  & 37.0 & 29.9 & 42.6   & 38.9 & 32.5 \\\midrule
    & & DO   & 0.87 & 0.58  & 0.13 & 0.85   & 0.62 & 0.09 & 0.87   & 0.66 & 0.11  \\
    &SSIM & DB  & 0.93 & 0.90  & 0.72 & 0.92   & 0.88 & 0.70 & 0.93   & 0.90 & 0.75  \\
    & & DMB   & 0.97 & 0.94  & 0.86 & 0.96  & 0.95 & 0.88 & 0.98   & 0.94 & 0.91 \\
\midrule
    \end{tabular}
\end{adjustbox}
\vspace*{0.0in}
\label{tal:score-eval}
\end{table}

Table~\ref{tal:score-eval} gives the mean values of average IS, PSNR and SSIM across 100 reconstructed images over ImageNet for the various model architectures and feature inversion methods. We can observe three trends:
\begin{itemize}
    \item For the same model-layer pair (i.e. each column), DM-based method (DMB) achieves the highest average IS/PSNR/SSIM compared to the Direct Optimization (DO) and Decoder-based (DB) approaches. This shows that our diffusion-based feature inversion attack is highly effective.
    \item For the same model, inverting features extracted from later layers results in lower IS/PSNR/SSIM for all three methods. However, with our diffusion-based method, the reduction in reconstruction quality is much less pronounced, showing that the attack is more capable of inverting later layer features. 
    \item In comparison, inverting features for ViT models is more challenging, although it remains feasible to invert features using output from middle layers of ViT (e.g., layer 5).
\end{itemize}

\subsection{Feature Inversion Results over YOLO}
\label{sec:white-box-eval-object-detection}
\begin{figure}
    \centering \includegraphics[width=0.48\textwidth]{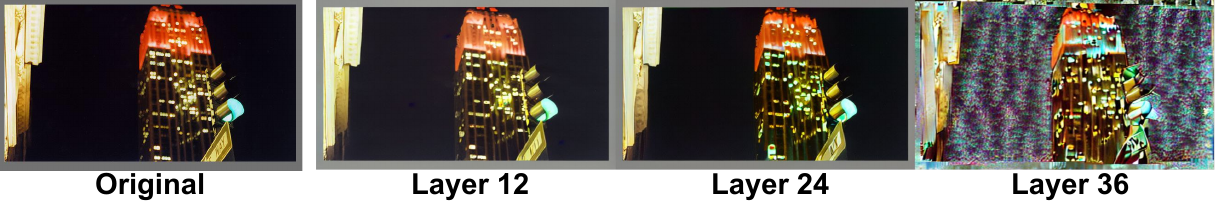}
    \caption{White-box feature inversion from different layers of YOLO.}
    \label{fig:yolo-rebuttal}
\end{figure}
We present additional feature inversion results over a YOLO-v2 model based on ResNet-50 for object detection (Figure~\ref{fig:yolo-rebuttal}). We select 100 test datasets from the COCO dataset for inversion. Specifically, for YOLO, our method can achieve higher average inception scores of 8.13, 7.22, and 6.32 using features from layers 12, 24, and 36. This is much higher than IS scores obtained by DO method (5.60, 3.75 and 1.29 for L12, L24 and L36, respectively) and DB method (6.96, 5.80 and 4.47 for L12, L24 and L36, respectively). 

\subsection{End-to-end Feature Inversion over CLIP}
\label{sec:white-box-clip-inverseion}

\begin{figure}
    \centering \includegraphics[width=0.4\textwidth]{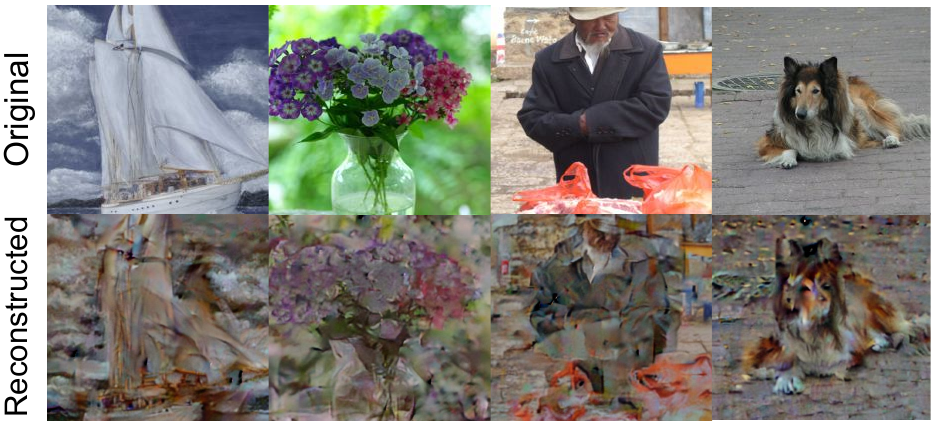}
    \caption{End-to-end feature inversion over CLIP image encoder.}
    \label{fig:white-clip-inversion}
    \vspace*{-0.17in}
\end{figure}

In this section, we present the results for the~\textbf{end-to-end feature inversion} over the CLIP~\cite{radford2021learning} image encoder. CLIP is language-visual multimodal DNN capable of understanding images and text jointly in a zero-shot manner, without the need for fine-tuning on a specific task. 
It aligns natural language prompts with images to perform a wide range of tasks, including image classification, image generation, and image-text retrieval. Specifically, the CLIP image encoder has been widely adopted for various of computer vision tasks including face recognition~\cite{bhat2023face,shen2023clip}, image segmentation~\cite{wang2022cris} and emotion classification~\cite{bondielli2021leveraging,deng2022learning}. We use the methodology described in Section~\ref{sec:white-box-feature-inversion} to invert the output features from the pretrained CLIP image encoder. Specifically, we download~\textit{clip-vit-base-patch32} from the huggingface official website~\cite{huggingface_clip}, and apply the settings described in the beginning of Section~\ref{sec:eval} for evaluation. We select 100 images from the ImageNet test dataset, some images together with their reconstructed versions are shown in Figure~\ref{fig:white-clip-inversion}. Specifically, we obtain an IS, PSNR and SSIM of 3.54, 13.2 and 0.50, respectively. In comparison, DO method achieves a IS, PSNR and SSIM of 0.88, 5.78 and 0.11, respectively. Similarly, the DB method also attains IS, PSNR, and SSIM scores of 2.21, 9.74 and 0.33, respectively. Furthermore, we note that, when compared with a DNN designed for image classification, CLIP is much easier to invert at the same layer depth. This enhancement could be attributed to CLIP's tendency to retain more original image data to support a wide range of downstream tasks, thereby aiding feature inversion. In contrast, DNNs for image classification typically discard redundant information, preserving only the essential data required for recognizing object classes within the image.

\subsection{Impact of Text Prior on Feature Inversion}
\label{sec:text-prior-eval}
In this section, we evaluate the impact of the text prior on the feature inversion results. Particularly, we extract intermediate features from deeper layers, such as layer 48 in ResNet-50. When attempting to invert these features from the deep layers, we observe a substantial quality degradation over the reconstructed input with the methods in Section~\ref{sec:white-box-feature-inversion}. This decline can be primarily attributed to the loss of critical information embedded within deep features.

\begin{figure}
    \centering
    \includegraphics[width=0.45\textwidth]{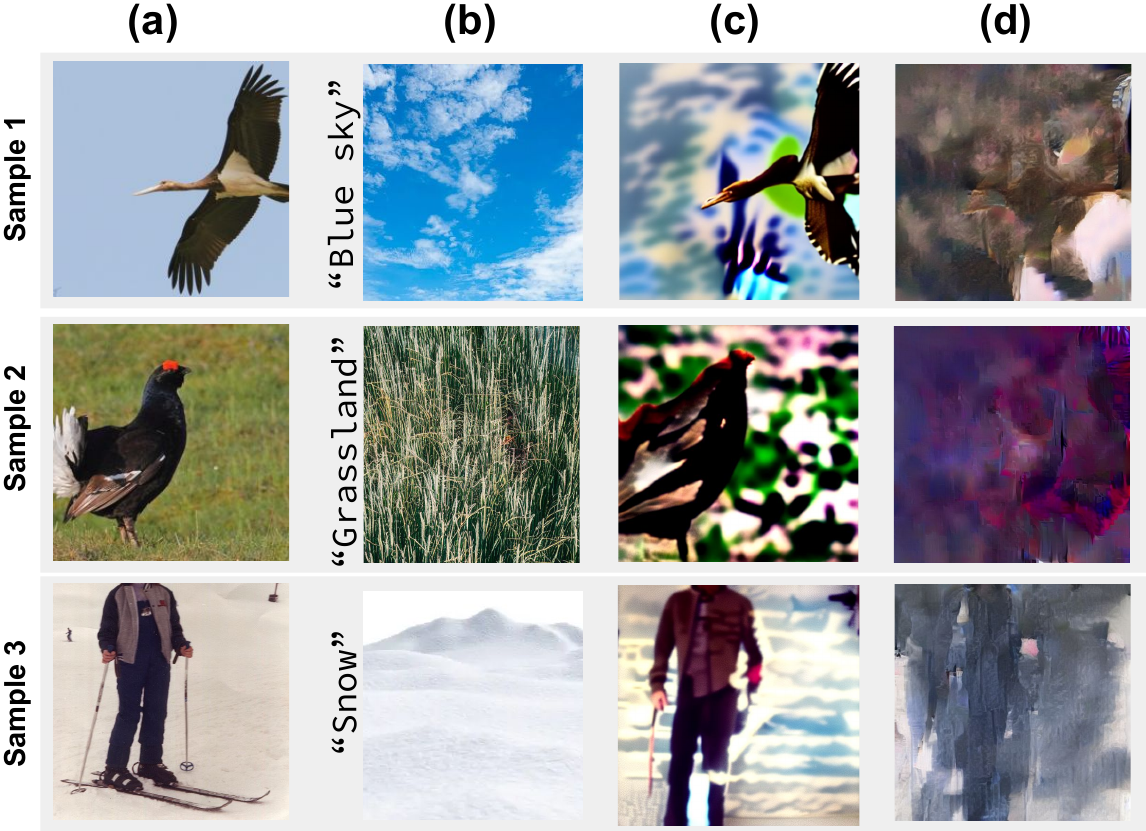}
    \caption{Impact of the text prior on feature inversion. Three samples with different scenes are selected from the ImageNet, shown in column (a). Column (b) shows the textual inputs and the corresponding synthetic images. Column (c) and (d) depict the reconstructed input with and without the text description.}
    \label{fig:text-prior-results}
\end{figure}

To evaluate the impact of textual prior, we select multiple images from the ImageNet dataset featuring simple backgrounds, including scenes with sky, grassland, and snow (Figure~\ref{fig:text-prior-results} (a)). For each of these three background categories, we choose 10 images. We utilize the textual descriptions ``blue sky'', ``a piece of grassland'', and ``snow'' as textual guidance for the LDMs during the feature inversion process. Subsequently, these selected images are forwarded through ResNet-18, ResNet-50 and ViT-based, and we extract the results from the output of layer 16 for ResNet-18, layer 48 for ResNet-50, and layer 8 for ViT-base, respectively. The extracted intermediate results are then employed to reconstruct the input images using the techniques outlined in Section~\ref{sec:White-box-text-prior}. Throughout the reconstruction process, $\lambda_{s}$ and $\lambda_{txt}$ in Equation~\ref{eqn:text-prior-loss} are set to 1 and 10, respectively. The rest of the settings remain the same as those in Section~\ref{sec:inversion-diffusion-eval}.

The reconstruction results are shown in Figure~\ref{fig:text-prior-results}.
We notice a significant enhancement in the feature inversion quality when incorporating the textual prior, as depicted in Figure~\ref{fig:text-prior-results} (c). These reconstructions, although not pixel-perfect, closely resemble the original images in (a) in terms of the semantic content. In comparison, the reconstructions obtained without textual prior, as shown in Figure~\ref{fig:text-prior-results} (d), are not semantically meaningful. Table~\ref{tal:text-prior-scores} shows quantitative results for feature inversion with (right) and without (left) textual prior in terms of average IS, PSNR and SSIM. Evidently, including the textual prior significantly improves the reconstruction quality quantitatively as well.
\begin{figure}
    \centering
    \includegraphics[width=0.4\textwidth]{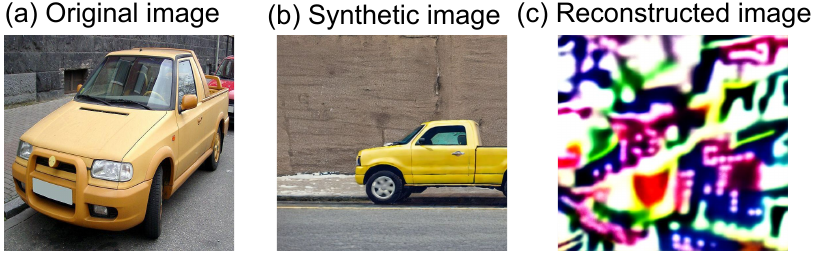}
    \caption{Limitations on utilizing text priors.}
    \label{fig:text-prior-findings}
\end{figure}
While the use of a textual prior can greatly enhance the reconstruction quality, it should be employed judiciously, as its improper use can potentially impair the reconstruction results. To illustrate this, we utilize intermediate results from layer 16 of ResNet-18 to reconstruct the user input shown in Figure~\ref{fig:text-prior-findings} (a), while providing the LDM with a text prior ``yellow pickup park along the road'' for feature inversion. Surprisingly, this does not lead to improved reconstruction quality, as depicted in Figure~\ref{fig:text-prior-findings} (c).
One possible explanation is this description fails to accurately characterize the object within the victim image, resulting in synthetic images that incorrectly represent the foreground in terms of shape, texture and position, as seen in Figure~\ref{fig:text-prior-findings} (b). This misalignment further degrades inversion quality. 
In general, we notice that offering a simple textual description of the image background tends to enhance the reconstruction performance. These descriptions provides an overview on the background of the image, outlining key attributes like the dominant color and surroundings. Our research marks the first step in investigating how textual descriptions influence feature inversion quality. Further investigation is needed to fully grasp the impact of textual priors.

\begin{figure}
    \centering
    \includegraphics[width=0.45\textwidth]{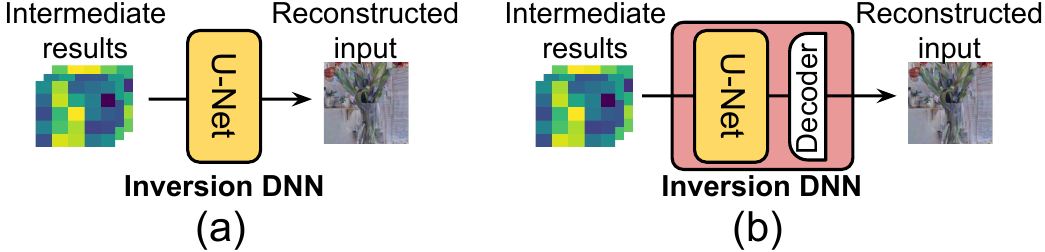}
    \caption{Baseline algorithms for black-box evaluation.}
    \label{fig:black-box-baselines}
\end{figure}

\begin{figure*}
    \centering
    \includegraphics[width=0.9\textwidth]{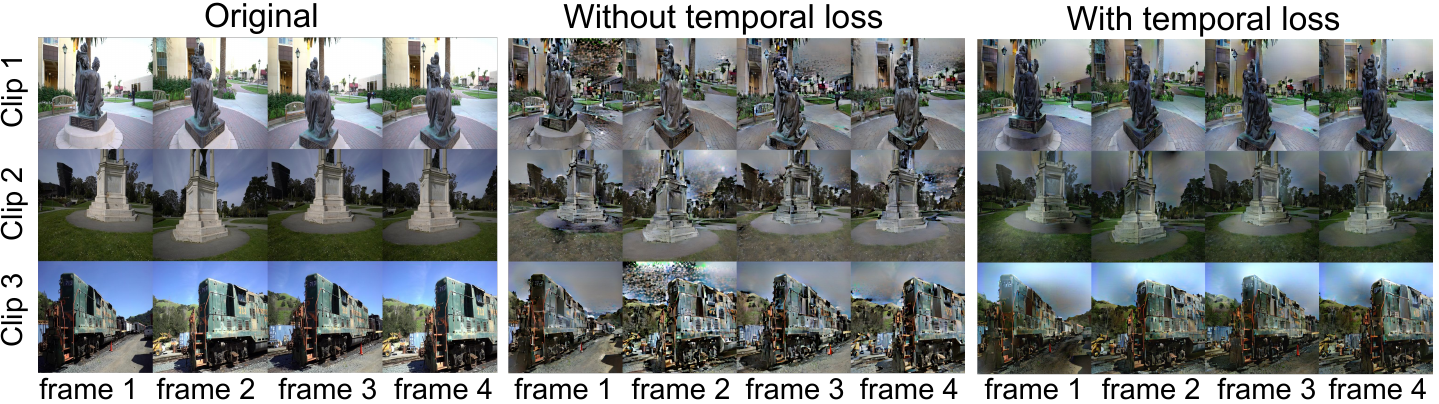}
    \caption{Feature inversion across multiple frames. Our method obtains average increase of 1.8, 7.3 and 0.39 for IS, PSNR and SSIM.}
    \label{fig:multi-frame-results}
\end{figure*}

\subsection{Multi-frame Feature Inversion}
\label{sec:multi-frame-results}
In this section, we evaluate the multi-frame feature inversion algorithm detailed in Section~\ref{sec:white-box-multi-frame}. To create multi-frame inputs with high correlations, we utilize the tanks and temples dataset~\cite{knapitsch2017}, which contains high-resolution video clips for twelve different objects. We select ten video clips and extract four consecutive frames from each video clip, the time interval between a pair of consecutive two frames is 0.5 seconds. Subsequently, we employ the loss function defined in Equation~\ref{eqn:temporal-loss} to jointly reconstruct the four frames. We configure $\lambda_{s}$ and $\lambda_{c}$ in Equation~\ref{eqn:temporal-loss} to be 1 and 5, respectively. To accelerate the reconstruction process, we set the sampling steps to 10. All other settings remain consistent with the earlier sections. To demonstrate the advance of the proposed loss function, we conduct a comparison of reconstructed image quality with and without the inclusion of the smoothing loss $\mathcal{L}_{c}$.

From the results presented in Figure~\ref{fig:multi-frame-results}, we observe a noticeably better reconstruction quality by using the smoothing loss. We also observe that the inclusion of the smoothing loss results in improvements in IS, PSNR, and SSIM scores, with an average increase of 1.8 for IS, 7.3 dB for PSNR, and 0.39 for SSIM across all the video clips.

\begin{table}
\caption{Feature inversion quality with and without textual priors. Left/right numbers show results without/with priors.}
\centering
\begin{adjustbox}{width=0.9\columnwidth,center}
\begin{tabular}{c|c|c|c} \hline
Metrics       &  ResNet-18 (L16)  & ResNet-50 (L48) & ViT-base (L8) \\ \hline
IS &   0.33/3.60  &   0.35/3.84 &  0.47/3.05 \\\hline
PSNR &  4.5/15.4  &  5.2/14.9  &  4.3/14.7 \\\hline
SSIM &   0.02/0.59   &  0.03/0.54  &  0.02/0.56\\\hline
\end{tabular}
\end{adjustbox}
\vspace*{0.0in}
\label{tal:text-prior-scores}
\end{table}

\subsection{Ablation Study on Number of Diffusion Sampling Steps}
\label{sec:ablation}
\begin{table}
\centering
\caption{IS and SSIM scores with different sampling steps.}
\begin{adjustbox}{width=0.9\columnwidth,center}
\begin{tabular}{c|ccccc } \hline
Metrics       &  10 steps  & 15 steps & 20 steps & 25 steps &  30 steps\\ \hline
IS &   5.93   &   6.38  &   6.55 &   6.62  &   6.65  \\\hline
SSIM &   0.80   &   0.87  &  0.88 &   0.89  &   0.89  \\\hline
\end{tabular}
\end{adjustbox}
\vspace*{-0.1in}
\label{tal:ablation-with-steps}
\end{table}
In this section, we investigate how the number of sampling steps affects the reconstruction quality. We utilize the intermediate features from layer 36 of ResNet-50 across 100 inputs from the ImageNet. Subsequently, we conduct feature inversion, as outlined in Section~\ref{sec:white-box-feature-inversion}, employing various sampling step values for the LDM. Table~\ref{tal:ablation-with-steps} depicts how the IS and SSIM scores evolve with different sampling steps. We observe that both scores increase as the number of sampling steps increases. Nevertheless, both stabilize when sampling steps exceeds 20. Therefore, in this study, we employ a sampling step value of 20 to achieve the optimal balance between feature inversion quality and training efficiency.

\section{Results for Black-box Feature Inversion}
\label{sec:black-eval}

\subsection{Experiment settings}
\label{sec:black-box-settings}
\textbf{Datasets and models:} We use the same datasets, target models and LDM as described in the Section~\ref{sec:white-box-settings}. To build the training dataset and test dataset of inversion DNN, we randomly select 4096 and 1024 images from the training and test datasets of either ImageNet or YOLO, respectively. We notice that a training data size of 4096 is enough for inversion DNN to generalize well.  

\textbf{Hyperparameters:} The inversion DNNs are trained over 96 epochs using a batch size of 128. We assign $\lambda_{s}$ values of 1 in equations~\ref{eqn:dm-black-loss},~\ref{eqn:dm-black-loss-text-prior}, and~\ref{eqn:black-multi-frame-loss}. We employ the Adam optimizer with an initial learning rate of 0.1 and $\beta$ values of (0.9, 0.999).

\textbf{Baseline:} Following the baseline setups outlined in~\ref{sec:white-box-settings}, we examine the black-box versions of DO and DB. In the case of DO (Figure~\ref{fig:black-box-baselines} (a)), we modify the architecture of the inversion DNN to directly reconstruct the user input $x$ without relying on the LDM. Conversely, for DB (Figure~\ref{fig:black-box-baselines} (b)), we integrate the decoder from the LDM into the inversion DNN to improve the quality of reconstruction. We change the structure of the inversion DNNs for DMB, DO and DB to ensure they contain an equal number of parameters.
DO has been applied by the~\cite{he2019model, dong2021privacy, maeng2022measuring} for feature inversion attack under the black-box setting, but DB has not been studied in prior works.

\begin{table}
\caption{Evaluation results for black-box inversion.}
 \centering
\begin{adjustbox}{width=\columnwidth,center}
    \begin{tabular}{lcccccccccccc}\toprule 
    & \multirow{2}{*}{\begin{tabular}{@{}c@{}}Metric\end{tabular} } & \multirow{2}{*}{\begin{tabular}[c]{@{}c@{}}Method\end{tabular} }
    & \multicolumn{3}{c}{ResNet-18} & \multicolumn{3}{c}{ResNet-50} & \multicolumn{3}{c}{ViT-base}  \\
    \cmidrule(lr){4-6} \cmidrule(lr){7-9} \cmidrule(lr){10-12}
    &  &  & L4  & L8  & L12  & L12  & L24  & L36  & L3  & L4  & L5 \\ \toprule
    & & DO   & 5.38 & 3.64  & 1.01 & 5.36  & 3.41 & 0.98 & 5.31 & 3.77 & 1.48 \\
    &IS & DB  & 6.53 & 5.23  & 3.60 & 6.79  & 5.20 & 3.21 & 6.62 & 5.16 & 3.51 \\
    & & DMB   & 6.99 & 6.21  & 5.19 & 7.08  & 6.44 & 4.89 & 6.98 & 6.27 & 5.00 \\\midrule
    & & DO   & 27.6 & 11.1  & 7.74 & 26.3   & 13.1 & 7.11 & 26.6  & 13.2 & 7.80 \\
    &PSNR & DB  & 34.3 & 24.3  & 9.6 & 33.4  & 27.0 & 12.3 & 35.3  & 23.9 & 11.5 \\
    & & DMB   & 40.4 & 32.5  & 20.6 & 39.2  & 31.4 & 23.9 & 41.6  & 31.5 & 19.5 \\\midrule
    & & DO   & 0.84 & 0.40  & 0.08 & 0.84   & 0.54 & 0.07 & 0.85 & 0.62 & 0.08  \\
    &SSIM & DB  & 0.90 & 0.78  & 0.38 & 0.92 & 0.79 & 0.43 & 0.91 & 0.80 & 0.55  \\
    & & DMB   & 0.92 & 0.84  & 0.46 & 0.94  & 0.88 & 0.67 & 0.95 & 0.88 & 0.71 \\
\midrule
    \end{tabular}
\end{adjustbox}
\vspace*{0.15in}
\label{tal:black-box-score-eval}
\vspace*{-0.2in}
\end{table}

\subsection{Feature Inversion of Split Models for Image Classification}
\label{sec:black-box-imagenet-evaluation}
\begin{figure}
    \centering
    \includegraphics[width=0.5\textwidth]{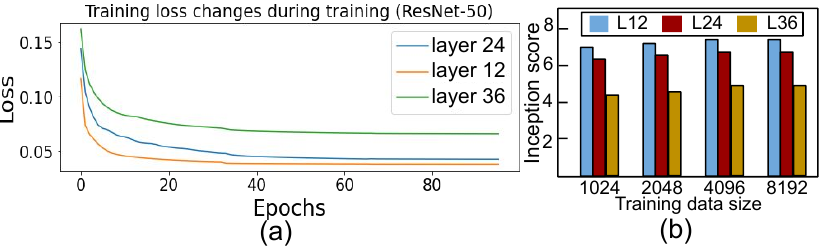}
    \caption{(a) Training loss vs. number of epochs for black-box feature inversion over ResNet-50 on ImageNet. Features are extracted from the end of layer 12, layer 24 and layer 36 respectively. (b) Inception scores with different training data size on ResNet-50 with ImageNet.}
    \label{fig:black-box-training-loss}
\end{figure}

Similar to Section~\ref{sec:inversion-diffusion-eval}, we begin by evaluating the reconstruction accuracy of our black-box feature inversion attacks \emph{without} utilizing text priors. For every target DNN model, we partition them into blocks of layers and capture intermediate outputs at the conclusion of each block. Next, we apply the methods presented in Section~\ref{sec:black-box-method} to reconstruct the user input. A sample training loss curve for inversion DNN is shown in Figure~\ref{fig:black-box-training-loss} (a).

Table~\ref{tal:black-box-score-eval} presents the mean IS, PSNR, and SSIM values computed across the test dataset for different model architectures and feature inversion techniques. The comparison on visual outputs is shown in the appendix. Notably, our approach (DM-based) consistently exhibits a higher reconstruction quality across all datasets and DNN architectures. Furthermore, it is worth noting that, under identical settings, the reconstruction quality tends to be lower for the black-box feature inversion attacks compared to the white-box feature inversion attacks, as shown in Section~\ref{sec:inversion-diffusion-eval}.

\subsection{Additional Results over YOLO and CLIP}
\label{sec:black-box-yolo-clip-inversion}

\begin{figure}
    \centering \includegraphics[width=0.48\textwidth]{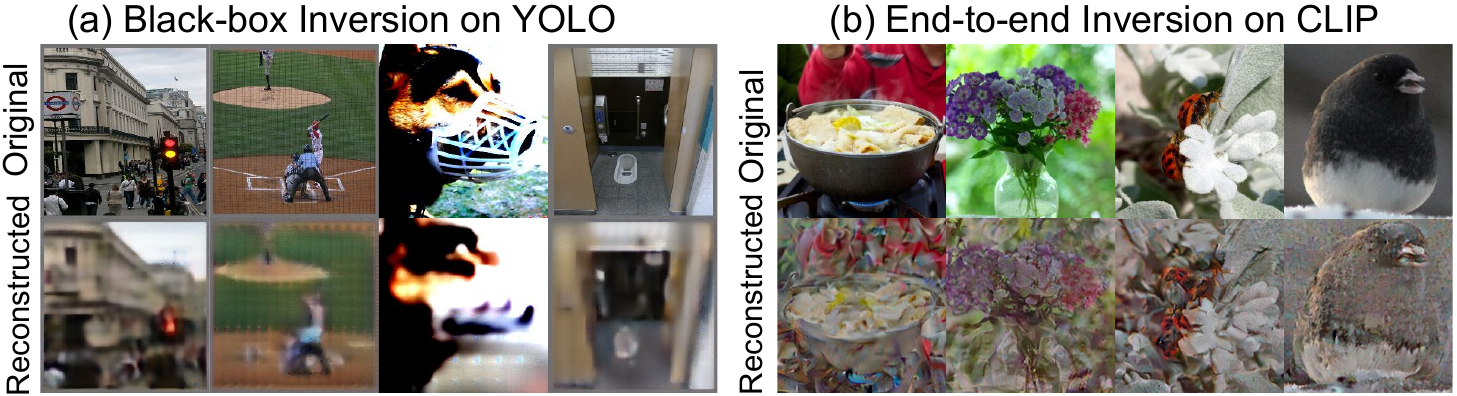}
    \caption{(a) Black-box feature inversion using the output from layer 36 of YOLO. (b) End-to-end black-box feature inversion attack over CLIP.}
    \label{fig:yolo-clip}
\end{figure}

We present additional results using a YOLO-v2 model based on ResNet-50 for object detection. The inputs are reconstructed by the inversion DNN using the intermediate results from layer 12, 24, and 36 (Figure~\ref{fig:yolo-clip} (a)). DMB achieves inception scores of 7.54, 6.99, and 6.80 using features from layers 12, 24, and 36 over the test dataset of COCO. This is much higher than IS scores obtained by DO method (5.64, 3.97 and 1.20 for L12, L24 and L36, respectively) and DB method (6.95, 5.95 and 4.46 for L12, L24 and L36, respectively).

Moreover, in Figure~\ref{fig:yolo-clip} (b), we showcase the outcomes of black-box end-to-end feature inversion using the CLIP~\cite{radford2021learning} image encoder. Specifically, we curate a subset of 1024 images from the ImageNet test dataset and achieve an IS, PSNR, and SSIM of 3.22, 12.0, and 0.46, respectively. Contrasting this, the DO method yields an IS, PSNR, and SSIM of 0.45, 2.62, and 0.09, respectively, while the DB method returns IS, PSNR, and SSIM scores of 2.03, 9.10, and 0.25, correspondingly. Similarly, we observe that CLIP is notably more amenable to inversion at the same layer depth compared to a DNN tailored for image classification.

\subsection{Impact of Text Prior on Feature Inversion}
\label{sec:black-box-text-prior-eval}
In this section, we analyze the impact of textual prior on the inversion results. Especially, we extract the intermediate feature from layer 16 in ResNet-18 and layer 48 from YOLO, 
we extract intermediate features from deeper layers, such as layer 48 in ResNet-50, and invert the image with simple background, as described in Section~\ref{sec:text-prior-eval}. $\lambda_{s}$ and $\lambda_{txt}$ in Equation~\ref{eqn:text-prior-loss} are set to 1 and 3, respectively. 

Table~\ref{tal:black-box-text-prior-scores} highlights the quantitative results for feature inversion with and without textual prior in terms of average IS, PSNR and SSIM. Evidently, including the textual prior significantly improves reconstruction quality quantitatively as well. However, we also observe that a similar failure case as described in Section~\ref{sec:text-prior-eval} under black-box setting. 

\begin{table}
\centering
\caption{Black-box feature inversion results with/without textual priors. Left/right numbers show results without/with priors.}
\begin{adjustbox}{width=0.85\columnwidth,center}
\begin{tabular}{c|c|c|c} \hline
Metrics & ResNet-50 (L48) &  YOLO (L48)  & CLIP (End-to-end)\\ \hline
IS &   0.29/3.11  &   0.88/3.48 &  3.22/4.09 \\\hline
PSNR &  4.3/13.6  &  5.4/14.0  &  12.0/17.7 \\\hline
SSIM &   0.03/0.50   &  0.08/0.44  &  0.46/0.56\\\hline
\end{tabular}
\end{adjustbox}
\vspace*{0.2in}
\label{tal:black-box-text-prior-scores}
\vspace*{-0.3in}
\end{table}

\subsection{Multi-frame Feature Inversion}
\label{sec:black-box-multi-frame-results}
In this section, we evaluate the multi-frame feature inversion algorithm detailed in Section~\ref{sec:black-box-multi-frame}.
We employ the same training and testing datasets as described in~\ref{sec:multi-frame-results}. Specifically, the training and test datasets include 1024 and 256 video clips, respectively. $\lambda_{s}$ in Equation~\ref{eqn:black-multi-frame-loss} are set to 1. We conduct a comparison of reconstructed image quality with and without the inclusion of the pointwise convolution layer described in Figure~\ref{fig:inversion-dnn-multiframe}, whose primary function is to consider the temporal correlation during the reconstruction of input frames. Moreover, we adjust the structure of the DNN so that the total amount of parameters are the same for both scenarios. We note a significant enhancement in IS, PSNR, and SSIM scores with the inclusion of the pointwise convolutional layer, resulting in an average increase of 0.2 for IS, 4.0 dB for PSNR, and 0.09 for SSIM across all video clips in the test dataset using the intermediate results from layer 36 of ResNet-50 on ImageNet dataset. More evaluation results are presented in the appendix.

\subsection{Ablation Study on Training Set Size}
\label{sec:black-box-ablation}
We study the impact of training dataset size on reconstruction quality. We vary the size of the ImageNet training dataset while maintaining the test dataset size at 1024, and measure feature inversion outcomes using the black-box setting. Figure~\ref{fig:black-box-training-loss} (b) demonstrates that IS scores steadily increase as the training dataset size grows. Interestingly, even with a training data size of 1024, a notably high IS score is achieved, suggesting that a smaller training dataset can still facilitate effective generalization of the inversion DNN.

\section{Discussion}
\label{sec:discussion}
In this section, we summarize some findings we observe from the evaluation results (Section~\ref{sec:findings}). We then discuss the potential defense strategies in Section~\ref{sec:defense-feature-inversion}.

\subsection{Insights From the Evaluation Results}
\label{sec:findings}
\textbf{A deeper DNN does not guarantee privacy:} Based on the results outlined in Section~\ref{sec:inversion-diffusion-eval} and Section~\ref{sec:black-box-imagenet-evaluation}, it becomes evident that a deeper DNN does not inherently ensure privacy. For instance, as observed in Table~\ref{tal:score-eval} and Table~\ref{tal:black-box-score-eval}, the quality of reconstructed input using the intermediate features at the 12th layer of ResNet-18 is notably inferior to that of the 24th layer output of ResNet-50. This suggests that the absolute layer depth alone does not guarantee any privacy protection. Instead, what matters is the~\textbf{relative layer depth} within the DNN. For instance, the reconstructed quality using the outputs of a middle layer of ResNet-50 (e.g., 24) is approximately equal to that from the middle layer (e.g., 8) of ResNet-18.

\textbf{Transformer is harder to invert than CNN:} Another trend we notice is that transformers, such as ViT, exhibit better privacy protection capabilities compared to Convolutional Neural Networks (CNNs). As indicated by the results in Table~\ref{tal:score-eval} and Table~\ref{tal:black-box-score-eval}, the quality of reconstructed input using features from the middle layer (e.g., 5) of ViT-base is comparable to that obtained using features from later layers in ResNet-18 or ResNet-50. This may contribute to the fact that transformers with self-attention
mechanism inherently amounts to a low-pass filter~\cite{wang2022anti}, this will eliminate a lot of high-frequency information within the original input image, making the intermediate features harder to invert. 
By contrast, CNNs typically extract information across wide frequency ranges~\cite{yosinski2015understanding}, thereby retaining more essential information for feature inversion. Nevertheless, further studies are needed.

\textbf{Self-supervised pretrained backbone models are easier to invert:} The evaluation results shown in Section~\ref{sec:white-box-clip-inverseion} and Section~\ref{sec:black-box-yolo-clip-inversion} illustrate that pretrained backbone models with self-supervised learning, are more amenable to inversion compared to DNNs trained with supervised learning frameworks tailored for specific tasks like image classification. This is due to the fact that the SSL-pretrained backbones tend to preserve a rich set of information that can be beneficial for various downstream tasks. The pretraining process typically involves learning representations that capture meaningful patterns and structures in the input data, which can generalize well to different tasks. In contrast, DNNs trained with supervised learning using labeled datasets tend to eliminate redundant information unrelated to the task during the training process, making the input data more challenging to reconstruct.

\textbf{A well-constructed textual prompt can improve inversion performance: }
This work is the first work to demonstrate that additional information in another format (i.e., textual format) can be utilized to enhance reconstruction performance. Our general finding is that a simple textual description consisting of a few words that provides an overview of the image background, highlighting attributes like the dominant color and surroundings, can generally enhance reconstruction quality. Although applying an inaccurate textual prior will degrade the attack quality, if the attacker has multiple candidate textual descriptions, the best strategy is to exhaustively try all of them and select the one that achieves the optimal quality. While using textual priors in feature inversion is not our main focus, it is a promising area for future research.

\textbf{Revealing the model weight and structure can improve the quality of the attack:} Based on the evaluation  results presented in Table~\ref{tal:score-eval} and Table~\ref{tal:black-box-score-eval}, it is evident that the feature inversion attack achieves higher quality results in the white-box setting compared to the black-box setting. This underscores the importance of weight and architecture of the user model in influencing the quality of feature inversion.

\subsection{Defense over Feature Inversion Attack}
\label{sec:defense-feature-inversion}

In this section, we explore potential defense strategies to mitigate the feature inversion attacks outlined in Section~\ref{sec:white-box-method} and Section~\ref{sec:black-box-method}. 
One defensive strategy is to ensure that all DNN computation and communication happen over encrypted data, i.e. through cryptographic methods like secure multiparty computation (MPC) or homomorphic encryption (HE). 

Secure MPC allows a group of $n$ untrusting parties to collaboratively compute a public function $f(x_1, x_2, ..., x_n)$ over their private inputs $x_1, x_2, ..., x_n$ without revealing any of their secret information. If the MPC scheme is expressive enough to implement large and complex neural networks like diffusion models, the implementation often has prohibitively large communication overheads and high computational complexity. For example, recent work on MPC implementation of VGG16 in the WAN setting leads to 37s latency (and training can take several weeks)~\cite{wagh2020falcon}. 

Similarly, HE schemes allow certain operations, such as arithmetic or boolean functions, to be applied to the ciphertext, thereby allowing privacy-preserving neural network evaluations without revealing sensitive information in plaintext form. But computing on ciphertexts over plaintext means both higher communication and computation costs. Additionally, HE requires additional computation steps like noise-management via bootstrapping. Prior implementations show significant slowdown, ~300s for encryption, DNN application, and decryption (up to 30s for a 5x5 convolutional layer to a simple 5-layer MNIST network, and 127s for pooling)~\cite{gilad2016cryptonets}.   
While these are promising and active research areas, at this time these methods are not widely deployed due to large overheads over an already computationally intense DNN architecture. 

Another feasible approach involves the use of differential privacy (DP) to introduce noise and obscure sensitive information. 
In particular, random noise $\epsilon$ can be directly integrated into the intermediate results $z_{mid}$, with the magnitude of the noise being controlled to achieve the desired level of DP. Nevertheless, it's crucial to acknowledge the trade-off between usability and privacy. When higher levels of noise are introduced,  there will be a corresponding drop in model accuracy. To mitigate this loss in accuracy, it is beneficial to take into account the influence of the injected noise during the training phase for the target DNN $F_{\theta}$(.).



\section{Conclusion}
\label{sec:conclusion}

In this study, we demonstrate the significant performance enhancement achievable in the feature inversion process via the utilization of the diffusion model. 
We also highlight the potential for utilizing diverse forms of prior knowledge, such as textual information and cross-frame correlations, to further improve the reconstruction quality. From the evaluation results, we show that GenAI, with its remarkable ability to synthesize realistic and coherent data, can also be utilized to detrimentally affect individuals' lives, particularly in the context of privacy breaches. This opens up interesting future avenues in a promising direction of research.
\newpage
\bibliographystyle{IEEEtranS}
\bibliography{refs}

\begin{thebibliography}{10}
\providecommand{\url}[1]{#1}
\csname url@samestyle\endcsname
\providecommand{\newblock}{\relax}
\providecommand{\bibinfo}[2]{#2}
\providecommand{\BIBentrySTDinterwordspacing}{\spaceskip=0pt\relax}
\providecommand{\BIBentryALTinterwordstretchfactor}{4}
\providecommand{\BIBentryALTinterwordspacing}{\spaceskip=\fontdimen2\font plus
\BIBentryALTinterwordstretchfactor\fontdimen3\font minus \fontdimen4\font\relax}
\providecommand{\BIBforeignlanguage}[2]{{%
\expandafter\ifx\csname l@#1\endcsname\relax
\typeout{** WARNING: IEEEtranS.bst: No hyphenation pattern has been}%
\typeout{** loaded for the language `#1'. Using the pattern for}%
\typeout{** the default language instead.}%
\else
\language=\csname l@#1\endcsname
\fi
#2}}
\providecommand{\BIBdecl}{\relax}
\BIBdecl

\bibitem{azure_face}
``Azure face recognition.'' 2018, https://azure. microsoft.com/en-us/services/cognitive-services/face/.

\bibitem{pysyft}
``Pysyft,'' 2020, https://github.com/OpenMined/PySyft.

\bibitem{splitnn}
``Splitnn,'' 2020, https://blog.openmined.org/tag/splitnn/.

\bibitem{amazon_face}
``Amazon rekognition.'' 2021, https://aws.amazon.com/rekognition.

\bibitem{huggingface_clip}
``Clip image encoder,'' 2021, https://huggingface.co/openai/clip-vit-base-patch32.

\bibitem{aggarwal2021fedface}
D.~Aggarwal, J.~Zhou, and A.~K. Jain, ``Fedface: Collaborative learning of face recognition model,'' in \emph{2021 IEEE International Joint Conference on Biometrics (IJCB)}.\hskip 1em plus 0.5em minus 0.4em\relax IEEE, 2021, pp. 1--8.

\bibitem{akintoye2023layer}
S.~B. Akintoye, L.~Han, H.~Lloyd, X.~Zhang, D.~Dancey, H.~Chen, and D.~Zhang, ``Layer-wise partitioning and merging for efficient and scalable deep learning,'' \emph{Future Generation Computer Systems}, vol. 149, pp. 432--444, 2023.

\bibitem{bhat2023face}
A.~Bhat and S.~Jain, ``Face recognition in the age of clip \& billion image datasets,'' \emph{arXiv preprint arXiv:2301.07315}, 2023.

\bibitem{bondielli2021leveraging}
A.~Bondielli, L.~C. Passaro \emph{et~al.}, ``Leveraging clip for image emotion recognition,'' in \emph{CEUR WORKSHOP PROCEEDINGS}, vol. 3015.\hskip 1em plus 0.5em minus 0.4em\relax CEUR-WS, 2021.

\bibitem{borgeaud2022improving}
S.~Borgeaud, A.~Mensch, J.~Hoffmann, T.~Cai, E.~Rutherford, K.~Millican, G.~B. Van Den~Driessche, J.-B. Lespiau, B.~Damoc, A.~Clark \emph{et~al.}, ``Improving language models by retrieving from trillions of tokens,'' in \emph{International conference on machine learning}.\hskip 1em plus 0.5em minus 0.4em\relax PMLR, 2022, pp. 2206--2240.

\bibitem{chu2020expressive}
H.~Chu, S.~Ma, F.~De~la Torre, S.~Fidler, and Y.~Sheikh, ``Expressive telepresence via modular codec avatars,'' in \emph{Computer Vision--ECCV 2020: 16th European Conference, Glasgow, UK, August 23--28, 2020, Proceedings, Part XII 16}.\hskip 1em plus 0.5em minus 0.4em\relax Springer, 2020, pp. 330--345.

\bibitem{deng2009imagenet}
J.~Deng, W.~Dong, R.~Socher, L.-J. Li, K.~Li, and L.~Fei-Fei, ``Imagenet: A large-scale hierarchical image database,'' in \emph{2009 IEEE conference on computer vision and pattern recognition}.\hskip 1em plus 0.5em minus 0.4em\relax Ieee, 2009, pp. 248--255.

\bibitem{deng2022learning}
S.~Deng, L.~Wu, G.~Shi, L.~Xing, M.~Jian, and Y.~Xiang, ``Learning to compose diversified prompts for image emotion classification,'' \emph{arXiv preprint arXiv:2201.10963}, 2022.

\bibitem{ding2023resource}
A.~Ding, A.~Hass, M.~Chan, N.~Sehatbakhsh, and S.~Zonouz, ``Resource-aware dnn partitioning for privacy-sensitive edge-cloud systems,'' in \emph{International Conference on Neural Information Processing}.\hskip 1em plus 0.5em minus 0.4em\relax Springer, 2023, pp. 188--201.

\bibitem{dong2022splitnets}
X.~Dong, B.~De~Salvo, M.~Li, C.~Liu, Z.~Qu, and Z.~Li, ``Splitnets: Designing neural architectures for efficient distributed computing on head-mounted systems,'' in \emph{Proceedings of the IEEE/CVF Conference on Computer Vision and Pattern Recognition}, 2022, pp. 12\,559--12\,569.

\bibitem{dong2021privacy}
X.~Dong, H.~Yin, J.~M. Alvarez, J.~Kautz, P.~Molchanov, and H.~Kung, ``Privacy vulnerability of split computing to data-free model inversion attacks,'' \emph{arXiv preprint arXiv:2107.06304}, 2021.

\bibitem{dosovitskiy2016inverting}
A.~Dosovitskiy and T.~Brox, ``Inverting visual representations with convolutional networks,'' in \emph{Proceedings of the IEEE conference on computer vision and pattern recognition}, 2016, pp. 4829--4837.

\bibitem{erdougan2022unsplit}
E.~Erdo{\u{g}}an, A.~K{\"u}p{\c{c}}{\"u}, and A.~E. {\c{C}}i{\c{c}}ek, ``Unsplit: Data-oblivious model inversion, model stealing, and label inference attacks against split learning,'' in \emph{Proceedings of the 21st Workshop on Privacy in the Electronic Society}, 2022, pp. 115--124.

\bibitem{feltin2023dnn}
T.~Feltin, L.~March{\'o}, J.-A. Cordero-Fuertes, F.~Brockners, and T.~H. Clausen, ``Dnn partitioning for inference throughput acceleration at the edge,'' \emph{IEEE Access}, vol.~11, pp. 52\,236--52\,249, 2023.

\bibitem{fu2023auto}
Y.~Fu, Y.~Li, C.~Li, J.~Saragih, P.~Zhang, X.~Dai, and Y.~C. Lin, ``Auto-card: Efficient and robust codec avatar driving for real-time mobile telepresence,'' in \emph{Proceedings of the IEEE/CVF Conference on Computer Vision and Pattern Recognition}, 2023, pp. 21\,036--21\,045.

\bibitem{gilad2016cryptonets}
R.~Gilad-Bachrach, N.~Dowlin, K.~Laine, K.~Lauter, M.~Naehrig, and J.~Wernsing, ``Cryptonets: Applying neural networks to encrypted data with high throughput and accuracy,'' in \emph{International conference on machine learning}.\hskip 1em plus 0.5em minus 0.4em\relax PMLR, 2016, pp. 201--210.

\bibitem{hauswald2014hybrid}
J.~Hauswald, T.~Manville, Q.~Zheng, R.~Dreslinski, C.~Chakrabarti, and T.~Mudge, ``A hybrid approach to offloading mobile image classification,'' in \emph{2014 IEEE International Conference on Acoustics, Speech and Signal Processing (ICASSP)}.\hskip 1em plus 0.5em minus 0.4em\relax IEEE, 2014, pp. 8375--8379.

\bibitem{he2019model}
Z.~He, T.~Zhang, and R.~B. Lee, ``Model inversion attacks against collaborative inference,'' in \emph{Proceedings of the 35th Annual Computer Security Applications Conference}, 2019, pp. 148--162.

\bibitem{ho2020denoising}
J.~Ho, A.~Jain, and P.~Abbeel, ``Denoising diffusion probabilistic models,'' \emph{Advances in neural information processing systems}, vol.~33, pp. 6840--6851, 2020.

\bibitem{hyvarinen2000independent}
A.~Hyv{\"a}rinen and E.~Oja, ``Independent component analysis: algorithms and applications,'' \emph{Neural networks}, vol.~13, no. 4-5, pp. 411--430, 2000.

\bibitem{jeon2021gradient}
J.~Jeon, K.~Lee, S.~Oh, J.~Ok \emph{et~al.}, ``Gradient inversion with generative image prior,'' \emph{Advances in neural information processing systems}, vol.~34, pp. 29\,898--29\,908, 2021.

\bibitem{kang2022dnn}
W.~Kang, S.~Chung, J.~Y. Kim, Y.~Lee, K.~Lee, J.~Lee, K.~G. Shin, and H.~S. Chwa, ``Dnn-sam: Split-and-merge dnn execution for real-time object detection,'' in \emph{2022 IEEE 28th Real-Time and Embedded Technology and Applications Symposium (RTAS)}.\hskip 1em plus 0.5em minus 0.4em\relax IEEE, 2022, pp. 160--172.

\bibitem{kang2017neurosurgeon}
Y.~Kang, J.~Hauswald, C.~Gao, A.~Rovinski, T.~Mudge, J.~Mars, and L.~Tang, ``Neurosurgeon: Collaborative intelligence between the cloud and mobile edge,'' \emph{ACM SIGARCH Computer Architecture News}, vol.~45, no.~1, pp. 615--629, 2017.

\bibitem{karjee2022split}
J.~Karjee, P.~Naik, K.~Anand, and V.~N. Bhargav, ``Split computing: Dnn inference partition with load balancing in iot-edge platform for beyond 5g,'' \emph{Measurement: Sensors}, vol.~23, p. 100409, 2022.

\bibitem{knapitsch2017}
A.~Knapitsch, J.~Park, Q.-Y. Zhou, and V.~Koltun, ``Tanks and temples: Benchmarking large-scale scene reconstruction,'' \emph{ACM Transactions on Graphics}, vol.~36, no.~4, 2017.

\bibitem{lee2023wireless}
J.~Lee, H.~Lee, and W.~Choi, ``Wireless channel adaptive dnn split inference for resource-constrained edge devices,'' \emph{IEEE Communications Letters}, 2023.

\bibitem{lezama2017not}
J.~Lezama, Q.~Qiu, and G.~Sapiro, ``Not afraid of the dark: Nir-vis face recognition via cross-spectral hallucination and low-rank embedding,'' in \emph{Proceedings of the IEEE conference on computer vision and pattern recognition}, 2017, pp. 6628--6637.

\bibitem{Li_2022_CVPR}
Z.~Li, J.~Zhang, L.~Liu, and J.~Liu, ``Auditing privacy defenses in federated learning via generative gradient leakage,'' in \emph{Proceedings of the IEEE/CVF Conference on Computer Vision and Pattern Recognition (CVPR)}, June 2022, pp. 10\,132--10\,142.

\bibitem{lin2014microsoft}
T.-Y. Lin, M.~Maire, S.~Belongie, J.~Hays, P.~Perona, D.~Ramanan, P.~Doll{\'a}r, and C.~L. Zitnick, ``Microsoft coco: Common objects in context,'' in \emph{Computer Vision--ECCV 2014: 13th European Conference, Zurich, Switzerland, September 6-12, 2014, Proceedings, Part V 13}.\hskip 1em plus 0.5em minus 0.4em\relax Springer, 2014, pp. 740--755.

\bibitem{lu2022dpm}
C.~Lu, Y.~Zhou, F.~Bao, J.~Chen, C.~Li, and J.~Zhu, ``Dpm-solver: A fast ode solver for diffusion probabilistic model sampling in around 10 steps,'' \emph{Advances in Neural Information Processing Systems}, vol.~35, pp. 5775--5787, 2022.

\bibitem{lu2020deep}
H.~Lu, M.~Zhang, X.~Xu, Y.~Li, and H.~T. Shen, ``Deep fuzzy hashing network for efficient image retrieval,'' \emph{IEEE transactions on fuzzy systems}, vol.~29, no.~1, pp. 166--176, 2020.

\bibitem{lu2017hierarchical}
X.~Lu, Y.~Chen, and X.~Li, ``Hierarchical recurrent neural hashing for image retrieval with hierarchical convolutional features,'' \emph{IEEE transactions on image processing}, vol.~27, no.~1, pp. 106--120, 2017.

\bibitem{luo2023split}
D.~Luo, T.~Yu, Y.~Wu, H.~Wu, T.~Wang, and W.~Zhang, ``Split: Qos-aware dnn inference on shared gpu via evenly-sized model splitting,'' in \emph{Proceedings of the 52nd International Conference on Parallel Processing}, 2023, pp. 605--614.

\bibitem{lyu2022accelerating}
Z.~Lyu, X.~Xu, C.~Yang, D.~Lin, and B.~Dai, ``Accelerating diffusion models via early stop of the diffusion process,'' \emph{arXiv preprint arXiv:2205.12524}, 2022.

\bibitem{ma2021pixel}
S.~Ma, T.~Simon, J.~Saragih, D.~Wang, Y.~Li, F.~De~La~Torre, and Y.~Sheikh, ``Pixel codec avatars,'' in \emph{Proceedings of the IEEE/CVF Conference on Computer Vision and Pattern Recognition}, 2021, pp. 64--73.

\bibitem{maeng2022measuring}
K.~Maeng, C.~Guo, S.~Kariyappa, and E.~Suh, ``Measuring and controlling split layer privacy leakage using fisher information,'' \emph{arXiv preprint arXiv:2209.10119}, 2022.

\bibitem{mahendran2015understanding}
A.~Mahendran and A.~Vedaldi, ``Understanding deep image representations by inverting them,'' in \emph{Proceedings of the IEEE conference on computer vision and pattern recognition}, 2015, pp. 5188--5196.

\bibitem{mai2018reconstruction}
G.~Mai, K.~Cao, P.~C. Yuen, and A.~K. Jain, ``On the reconstruction of face images from deep face templates,'' \emph{IEEE transactions on pattern analysis and machine intelligence}, vol.~41, no.~5, pp. 1188--1202, 2018.

\bibitem{matsubara2019distilled}
Y.~Matsubara, S.~Baidya, D.~Callegaro, M.~Levorato, and S.~Singh, ``Distilled split deep neural networks for edge-assisted real-time systems,'' in \emph{Proceedings of the 2019 Workshop on Hot Topics in Video Analytics and Intelligent Edges}, 2019, pp. 21--26.

\bibitem{matsubara2022split}
Y.~Matsubara, M.~Levorato, and F.~Restuccia, ``Split computing and early exiting for deep learning applications: Survey and research challenges,'' \emph{ACM Computing Surveys}, vol.~55, no.~5, pp. 1--30, 2022.

\bibitem{morris2023text}
J.~X. Morris, V.~Kuleshov, V.~Shmatikov, and A.~M. Rush, ``Text embeddings reveal (almost) as much as text,'' \emph{arXiv preprint arXiv:2310.06816}, 2023.

\bibitem{mubark2024asap}
W.~H. Mubark, J.~G. Kasula, and M.~Y. Sarwar~Uddin, ``Asap: Asynchronous split inference for accelerated dnn execution,'' in \emph{Proceedings of the 25th International Conference on Distributed Computing and Networking}, 2024, pp. 32--44.

\bibitem{park2022quiltnet}
J.~Park, H.~Kwon, S.~Kim, J.~Lee, M.~Ha, E.~Lim, M.~Imani, and Y.~Kim, ``Quiltnet: efficient deep learning inference on multi-chip accelerators using model partitioning,'' in \emph{Proceedings of the 59th ACM/IEEE Design Automation Conference}, 2022, pp. 1159--1164.

\bibitem{radford2021learning}
A.~Radford, J.~W. Kim, C.~Hallacy, A.~Ramesh, G.~Goh, S.~Agarwal, G.~Sastry, A.~Askell, P.~Mishkin, J.~Clark \emph{et~al.}, ``Learning transferable visual models from natural language supervision,'' in \emph{International conference on machine learning}.\hskip 1em plus 0.5em minus 0.4em\relax PMLR, 2021, pp. 8748--8763.

\bibitem{reed2016generative}
S.~Reed, Z.~Akata, X.~Yan, L.~Logeswaran, B.~Schiele, and H.~Lee, ``Generative adversarial text to image synthesis,'' in \emph{International conference on machine learning}.\hskip 1em plus 0.5em minus 0.4em\relax PMLR, 2016, pp. 1060--1069.

\bibitem{richard2021audio}
A.~Richard, C.~Lea, S.~Ma, J.~Gall, F.~De~la Torre, and Y.~Sheikh, ``Audio-and gaze-driven facial animation of codec avatars,'' in \emph{Proceedings of the IEEE/CVF winter conference on applications of computer vision}, 2021, pp. 41--50.

\bibitem{rombach2022high}
R.~Rombach, A.~Blattmann, D.~Lorenz, P.~Esser, and B.~Ommer, ``High-resolution image synthesis with latent diffusion models,'' in \emph{Proceedings of the IEEE/CVF conference on computer vision and pattern recognition}, 2022, pp. 10\,684--10\,695.

\bibitem{rudin1992nonlinear}
L.~I. Rudin, S.~Osher, and E.~Fatemi, ``Nonlinear total variation based noise removal algorithms,'' \emph{Physica D: nonlinear phenomena}, vol.~60, no. 1-4, pp. 259--268, 1992.

\bibitem{saharia2022photorealistic}
C.~Saharia, W.~Chan, S.~Saxena, L.~Li, J.~Whang, E.~L. Denton, K.~Ghasemipour, R.~Gontijo~Lopes, B.~Karagol~Ayan, T.~Salimans \emph{et~al.}, ``Photorealistic text-to-image diffusion models with deep language understanding,'' \emph{Advances in Neural Information Processing Systems}, vol.~35, pp. 36\,479--36\,494, 2022.

\bibitem{salimans2016improved}
T.~Salimans, I.~Goodfellow, W.~Zaremba, V.~Cheung, A.~Radford, and X.~Chen, ``Improved techniques for training gans,'' \emph{Advances in neural information processing systems}, vol.~29, 2016.

\bibitem{schroff2015facenet}
F.~Schroff, D.~Kalenichenko, and J.~Philbin, ``Facenet: A unified embedding for face recognition and clustering,'' in \emph{Proceedings of the IEEE conference on computer vision and pattern recognition}, 2015, pp. 815--823.

\bibitem{shen2023clip}
S.~Shen, W.~Li, X.~Wang, D.~Zhang, Z.~Jin, J.~Zhou, and J.~Lu, ``Clip-cluster: Clip-guided attribute hallucination for face clustering,'' in \emph{Proceedings of the IEEE/CVF International Conference on Computer Vision}, 2023, pp. 20\,786--20\,795.

\bibitem{song2020information}
C.~Song and A.~Raghunathan, ``Information leakage in embedding models,'' in \emph{Proceedings of the 2020 ACM SIGSAC conference on computer and communications security}, 2020, pp. 377--390.

\bibitem{song2020denoising}
J.~Song, C.~Meng, and S.~Ermon, ``Denoising diffusion implicit models,'' \emph{arXiv preprint arXiv:2010.02502}, 2020.

\bibitem{song2020analyzing}
M.~Song, Z.~Wang, Z.~Zhang, Y.~Song, Q.~Wang, J.~Ren, and H.~Qi, ``Analyzing user-level privacy attack against federated learning,'' \emph{IEEE Journal on Selected Areas in Communications}, vol.~38, no.~10, pp. 2430--2444, 2020.

\bibitem{teerapittayanon2016branchynet}
S.~Teerapittayanon, B.~McDanel, and H.-T. Kung, ``Branchynet: Fast inference via early exiting from deep neural networks,'' in \emph{2016 23rd international conference on pattern recognition (ICPR)}.\hskip 1em plus 0.5em minus 0.4em\relax IEEE, 2016, pp. 2464--2469.

\bibitem{teerapittayanon2017distributed}
------, ``Distributed deep neural networks over the cloud, the edge and end devices,'' in \emph{2017 IEEE 37th international conference on distributed computing systems (ICDCS)}.\hskip 1em plus 0.5em minus 0.4em\relax IEEE, 2017, pp. 328--339.

\bibitem{wagh2020falcon}
S.~Wagh, S.~Tople, F.~Benhamouda, E.~Kushilevitz, P.~Mittal, and T.~Rabin, ``Falcon: Honest-majority maliciously secure framework for private deep learning,'' \emph{arXiv preprint arXiv:2004.02229}, 2020.

\bibitem{wang2022anti}
P.~Wang, W.~Zheng, T.~Chen, and Z.~Wang, ``Anti-oversmoothing in deep vision transformers via the fourier domain analysis: From theory to practice,'' \emph{arXiv preprint arXiv:2203.05962}, 2022.

\bibitem{wang2022cris}
Z.~Wang, Y.~Lu, Q.~Li, X.~Tao, Y.~Guo, M.~Gong, and T.~Liu, ``Cris: Clip-driven referring image segmentation,'' in \emph{Proceedings of the IEEE/CVF conference on computer vision and pattern recognition}, 2022, pp. 11\,686--11\,695.

\bibitem{wang2004image}
Z.~Wang, A.~C. Bovik, H.~R. Sheikh, and E.~P. Simoncelli, ``Image quality assessment: from error visibility to structural similarity,'' \emph{IEEE transactions on image processing}, vol.~13, no.~4, pp. 600--612, 2004.

\bibitem{xu2019ganobfuscator}
C.~Xu, J.~Ren, D.~Zhang, Y.~Zhang, Z.~Qin, and K.~Ren, ``Ganobfuscator: Mitigating information leakage under gan via differential privacy,'' \emph{IEEE Transactions on Information Forensics and Security}, vol.~14, no.~9, pp. 2358--2371, 2019.

\bibitem{yosinski2015understanding}
J.~Yosinski, J.~Clune, A.~Nguyen, T.~Fuchs, and H.~Lipson, ``Understanding neural networks through deep visualization,'' \emph{arXiv preprint arXiv:1506.06579}, 2015.

\bibitem{zeng2020coedge}
L.~Zeng, X.~Chen, Z.~Zhou, L.~Yang, and J.~Zhang, ``Coedge: Cooperative dnn inference with adaptive workload partitioning over heterogeneous edge devices,'' \emph{IEEE/ACM Transactions on Networking}, vol.~29, no.~2, pp. 595--608, 2020.

\bibitem{zhang2017stackgan}
H.~Zhang, T.~Xu, H.~Li, S.~Zhang, X.~Wang, X.~Huang, and D.~N. Metaxas, ``Stackgan: Text to photo-realistic image synthesis with stacked generative adversarial networks,'' in \emph{Proceedings of the IEEE international conference on computer vision}, 2017, pp. 5907--5915.

\bibitem{zhang2020adaptive}
S.~Q. Zhang, J.~Lin, and Q.~Zhang, ``Adaptive distributed convolutional neural network inference at the network edge with adcnn,'' in \emph{Proceedings of the 49th International Conference on Parallel Processing}, 2020, pp. 1--11.

\bibitem{zhang2021f}
X.~Zhang, D.~Wang, P.~Chuang, S.~Ma, D.~Chen, and Y.~Li, ``F-cad: A framework to explore hardware accelerators for codec avatar decoding,'' in \emph{2021 58th ACM/IEEE Design Automation Conference (DAC)}.\hskip 1em plus 0.5em minus 0.4em\relax IEEE, 2021, pp. 763--768.

\bibitem{Zhang_2020_CVPR}
Y.~Zhang, R.~Jia, H.~Pei, W.~Wang, B.~Li, and D.~Song, ``The secret revealer: Generative model-inversion attacks against deep neural networks,'' in \emph{Proceedings of the IEEE/CVF Conference on Computer Vision and Pattern Recognition (CVPR)}, June 2020.

\bibitem{zhou2017collaborative}
W.~Zhou, H.~Li, J.~Sun, and Q.~Tian, ``Collaborative index embedding for image retrieval,'' \emph{IEEE transactions on pattern analysis and machine intelligence}, vol.~40, no.~5, pp. 1154--1166, 2017.

\end{thebibliography}

\newpage
\newpage
\section{Appendix}
\label{sec:set-diff-dodis}
\subsection{statement on Data Availability}

Due to institutional restrictions, we are unable to use the public latent diffusion model for publishing our research outcomes. Instead, we employed an LDM with an architecture highly similar to Stable Diffusion 2.1 in terms of architecture, model size and pretraining techniques. Our internal model was pretrained on the dataset collected by a third party (Shutterstock) that is not public. Regarding the dataset, it consists of 385 million images: 321 million without people and 64 million with people.
In addition, we have also previously conducted extensive experiments with the public Stable Diffusion 2.1, which yielded similar (and even better) results than the reported results in terms of IS, PSNR, and SSIM scores.
If the paper is accepted, we intend to release the code as open source. This code will enable the integration of the public LDM for conducting feature inversion attacks.

\subsection{Implementation details}
Table~\ref{tab:white-box-petraining-setting} and Table~\ref{tab:black-box-petraining-setting} list the detailed settings for feature inversion described in Section~\ref{sec:white-box-method} and Section~\ref{sec:black-box-method}. To initiate the training process, the latent variable v is initialized using a randomly generated vector sampled from a normal distribution with a standard deviation of 0.1.
\begin{table}[h]
\centering
\begin{tabular}{l|l}
\toprule
General Configuration & Detail \\
\midrule
Optimizer & Adam \\
Total iterations & 1500 \\
Base learning rate & 0.1 \\
Learning rate schedule & multiple stages \\
Sampling steps & 20 \\
$\lambda_{s}$ & 1 \\
\bottomrule
\end{tabular}
\vspace*{0.15in}
\caption{Detailed settings for white-box feature inversion.}
\label{tab:white-box-petraining-setting}
\end{table}

\begin{figure}
    \centering
    \includegraphics[width=0.45\textwidth]{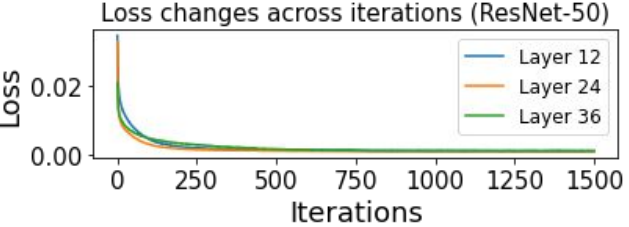}
    \caption{Changes on training loss during the inversion process.}
    \label{fig:loss-change}
\end{figure}

\begin{table}[h]
\centering
\begin{tabular}{l|l}
\toprule
General Configuration & Detail \\
\midrule
Optimizer & Adam \\
Total epochs & 96 \\
Base learning rate & 0.1 \\
Learning rate schedule & multiple stages \\
Sampling steps & 20 \\
Batch size & 128 \\
Training data size & 4096 \\
Test data size & 1024 \\
\bottomrule
\end{tabular}
\vspace*{0.15in}
\caption{Detailed settings for black-box feature inversion.}
\label{tab:black-box-petraining-setting}
\end{table}
In Figure~\ref{fig:loss-change}, we illustrate the changes on the loss function throughout the reconstruction process as outlined in Algorithm~\ref{alg:feature_inversion_alg} using the intermediate features from ResNet-50. It is evident that the loss values converge by the end of the 1500 iterations.

\subsection{Feature inversion training with textual prior}
Algorithm~\ref{alg:text_prior_alg} describes the algorithm for feature inversion training with text prior under white-box settings.
\begin{algorithm}
\caption{Feature Inversion with Text Prior}
\label{alg:text_prior_alg}
\DontPrintSemicolon
  \KwIn{
  $F_{1}(.)$ is the user DNN model. \\ 
  $v^{m}$ is the input latent vector of LDMs at iteration $m$. \\  
  $M$ is total number of iterations. \\
  $\epsilon$ is the learning rate. \\ 
  $t_{prior}$ is the prior knowledge described in text.\\
  $E()$ is the pretrained text encoder.\\
  } 
  \small
  \For{$1 \leq m \leq M$}{  
       {$v^{m}_{n} = \frac{v^{m}-\mathrm{mean}(v^{m})}{\mathrm{std}(v^{m})}$} \\
       {$\mathcal{L}_{tot} = ||F_{1}(D(v^{m}_{n}, E(t_{prior})))-z_{mid}||^{2}$ + $\lambda_{s}TV(D(v^{m}_{n}))$ + $\lambda_{txt} ||z_{n}-q||^{2}$}\\
       {$v^{m+1} = v^{m} - \epsilon \frac{d\mathcal{L}_{tot}}{dv}$} \\
       {$m = m + 1$} \\
   }
    {$v_{n} = \frac{v^{M}-\mathrm{mean}(v^{M})}{\mathrm{std}(v^{M})}$} \\
   {\Return $D(v_{n}^{M}, E(t_{prior}))$}.
\end{algorithm}

\begin{figure*}
    \centering
    \includegraphics[width=\textwidth]{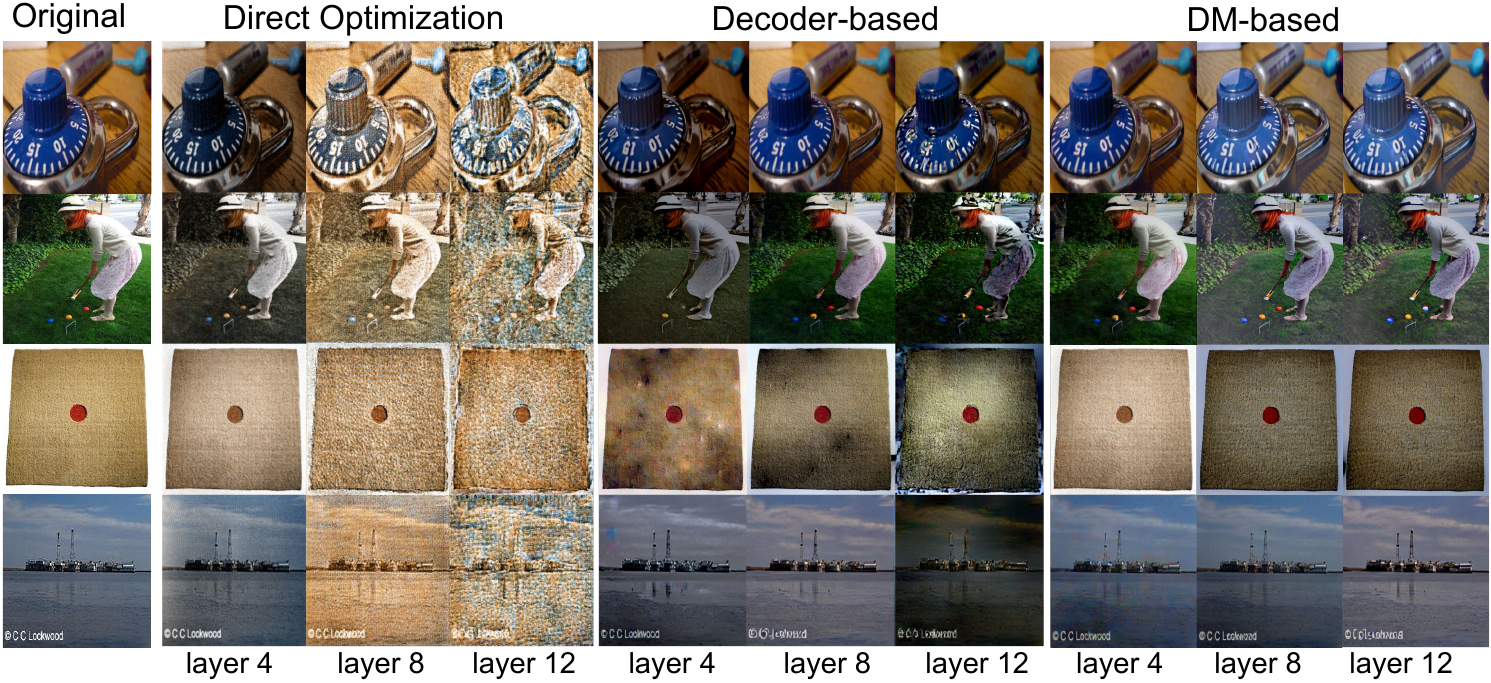}
    \caption{Feature inversion of ResNet-18 on ImageNet with white-box setting.}
    \label{fig:resnet18-imagenet}
\end{figure*}
\subsection{Multi-frame feature inversion training}
Algorithm~\ref{alg:multi_frame_alg} describes the algorithm for feature inversion training with multiple frames under white-box settings.

\subsection{Training algorithm for feature inversion under black-box settings}
Algorithm~\ref{alg:black_box_alg} describes the algorithm for feature inversion training with black-box setting.
\begin{algorithm}
\caption{Black-box Feature Inversion}
\label{alg:black_box_alg}
\DontPrintSemicolon
  \KwIn{
  $F_{1}(.)$ is the user DNN model. \\ 
  $X_{q} = \{x_{q}\}$, $Y_{q}= \{y_{q}\}$ are the sets training input samples and the corresponding intermediate results from the user DNN $F_{1}(.)$. \\  
  $(x_{q},y_{q})$ is a training sample. 
  $\lambda_{s}$ is the weight for the TV loss. \\
  $T$ is total number of iterations. \\
  $\epsilon$ is the learning rate. \\ 
  $F^{inv}_{\theta}(.)$ is the inversion DNN. \\ 
  $D(.)$ is the latent diffusion model. \\ 
  } 
  \small
  {Initialize $\theta$ within $F^{inv}_{\theta}(.)$.}\\
  \For{$1 \leq e \leq E$}{  
       \For{$(x_{q},y_{q})\in (X_{q},Y_{q})$}{
       {$z = F^{inv}_{\theta}(y_{q})$} \\
       {$x = D(z_{q})$} \\
       {$\mathcal{L}_{tot} = ||x-x_{q}||^{2}$ + $\lambda_{s}TV(x)$}\\
       {Compute the gradient and update $\theta$.} \\
       }
   }
   {\Return $\theta$}.
\end{algorithm}

\subsection{More results on white-box feature inversion}
\begin{figure*}
    \centering
    \includegraphics[width=0.94\textwidth]{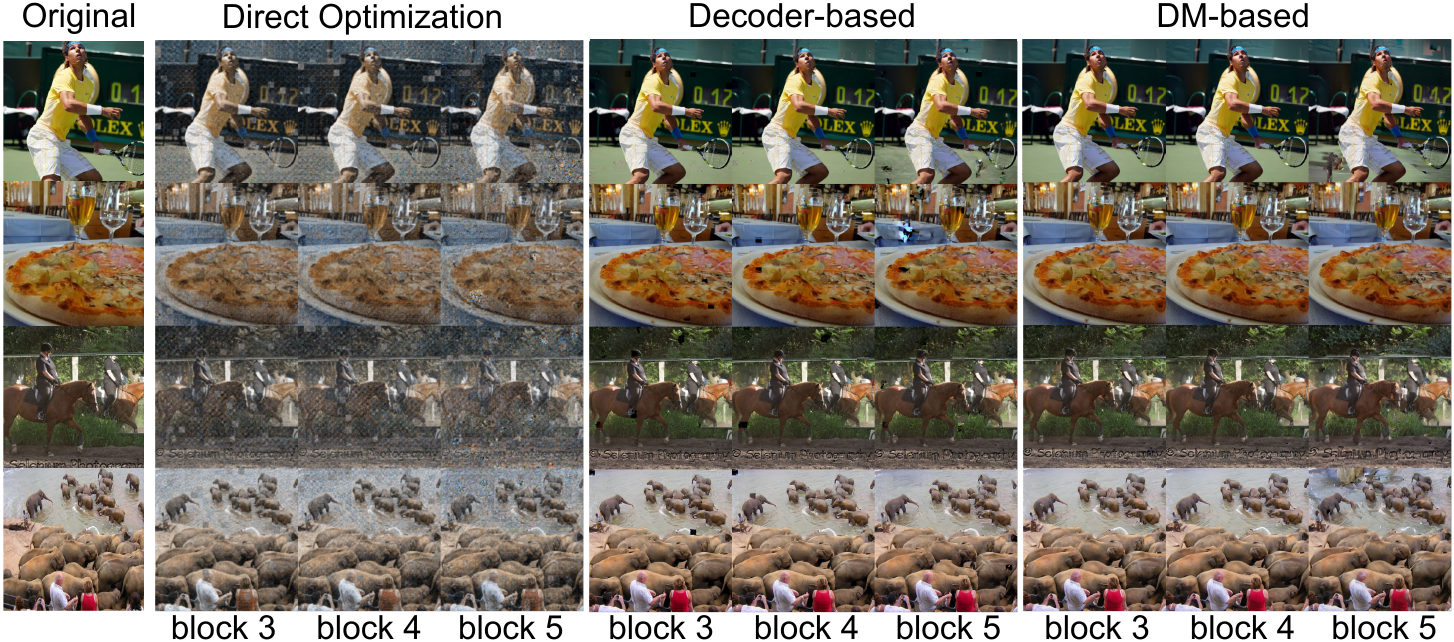}
    \caption{Feature inversion of ViT on ImageNet with white-box setting.}
    \label{fig:vit-imagenet}
\end{figure*}

Figure~\ref{fig:resnet18-imagenet} shows the feature inversion results for ResNet-18 on ImageNet dataset. Finally, Figure~\ref{fig:vit-imagenet} shows the feature inversion results for ViT on ImageNet.

\subsection{More results on black-box feature inversion}

\begin{figure*}
    \centering
    \includegraphics[width=0.94\textwidth]{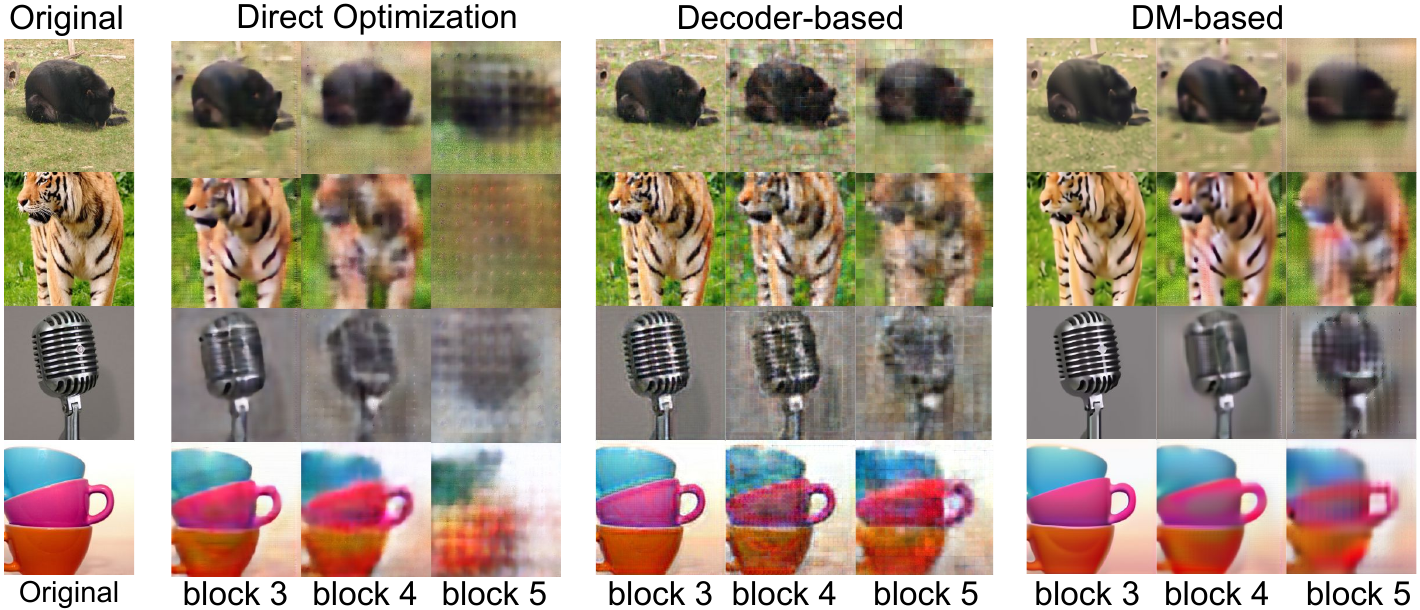}
    \caption{Feature inversion of ViT on ImageNet with black-box setting.}
    \label{fig:black-box-resnet18-imagenet}
\end{figure*}

Figure~\ref{fig:black-box-resnet18-imagenet} illustrate the feature inversion outcomes for ViT on ImageNet.

\subsection{Impact of sampling steps}
In this section, we show the impact of the DM sampling steps on the feature inversion results for white-box setting. Specifically, we show the reconstructed input images (Figure~\ref{fig:sampling-ablation}) from layer 36 of ResNet-50 on ImageNet under white-box settings. This serves as a supplementary addition to the ablation studies discussed in Section~\ref{sec:ablation}. We observe that the feature inversion quality improves as the number of sampling steps increasing from 10 to 20.

\subsection{Multi-frame inversion with black-box settings}
\label{sec:black-box-multi-frame}
In this section, we show the multi-frame feature inversion results under black-box settings (Table~\ref{tal:black-box-multi-frame}). We evaluate using two target models, ResNet-50 and ViT. For ResNet-50 and ViT, their features are inverted using the intermediate results from L36 and L5, respectively. We can see that involving the pointwise layer in the inversion DNN obtains a clear improvement on the reconstruction quality.
\begin{table}
\centering+
\caption{Multi-frame feature inversion results under black-box settings. For ResNet-50 and ViT, features are inverted using the outputs from L36 and L5, respectively. The number on the left/right represents the results obtained without/with a pointwise convolutional layer.}
\begin{adjustbox}{width=0.8\columnwidth,center}
\begin{tabular}{c|c|c|c} \hline
Metrics & IS &  PSNR  & SSIM\\ \hline
ResNet-50 &  4.82/4.99  &  23.6/27.2  &  0.67/0.76 \\\hline
ViT &   4.89/5.10   &  19.8/22.2  &  0.70/0.77\\\hline
\end{tabular}
\end{adjustbox}
\vspace*{0.17in}
\label{tal:black-box-multi-frame}
\end{table}

\subsection{More results on inversion with textual prior}
\begin{figure*}
    \centering
    \begin{minipage}{.5\textwidth}
        \centering
        \includegraphics[width=0.9\linewidth]{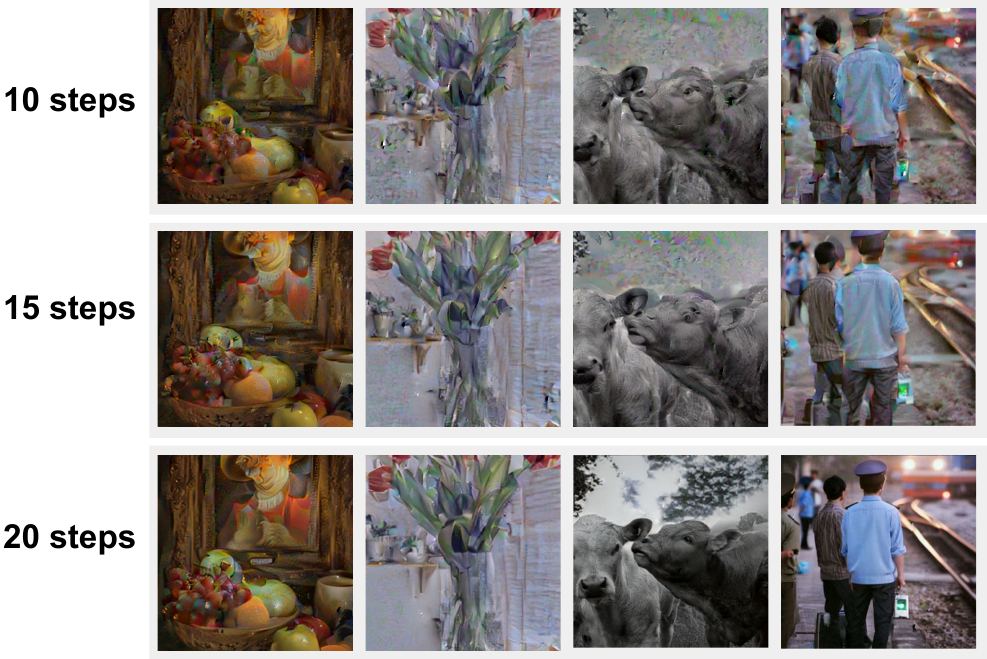}
        \caption{Feature inversion results with different sampling steps.}
        \label{fig:sampling-ablation}
    \end{minipage}%
    \begin{minipage}{0.5\textwidth}
        \centering
        \includegraphics[width=0.9\linewidth, height=0.6\linewidth]{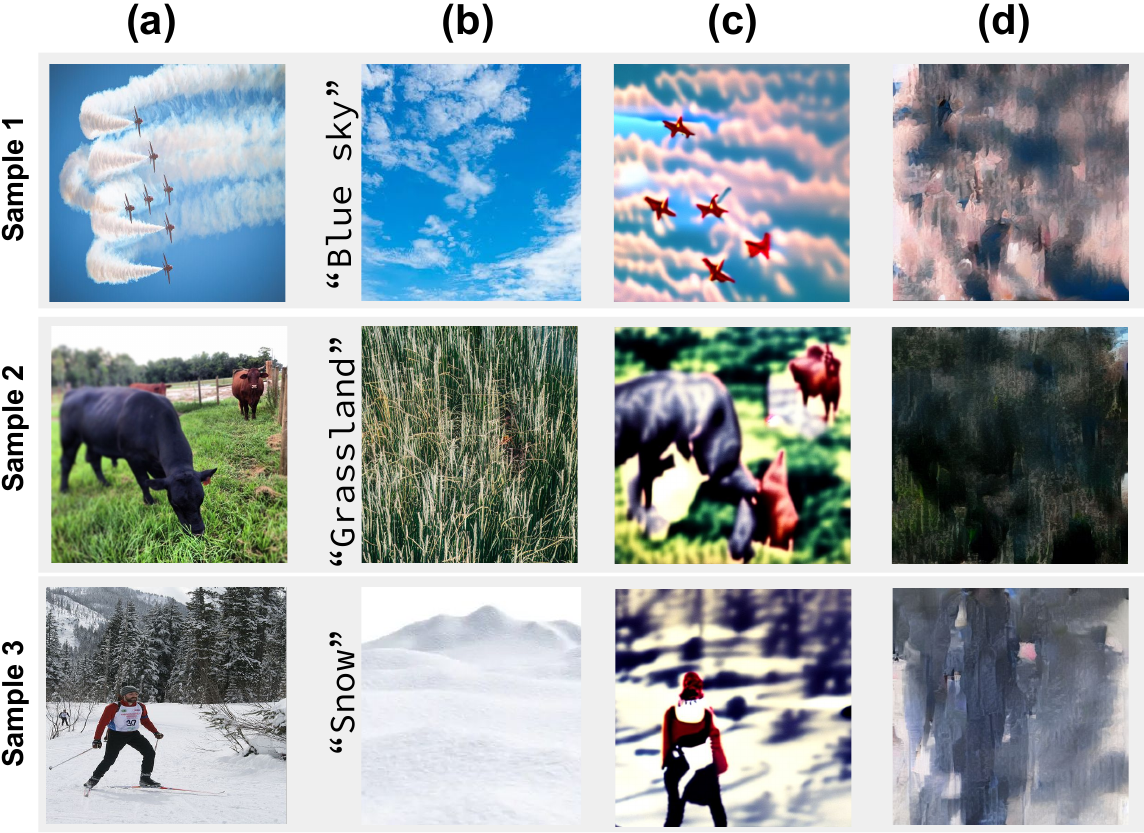}
        \caption{Feature inversion with textual prior.}
        \label{fig:new-txt-results}
    \end{minipage}
\end{figure*}

In this section, we show additional results on feature inversion with textual prior (Figure~\ref{fig:new-txt-results}) under white-box settings. Specifically, we use the same textual prior as Figure~\ref{fig:text-prior-results}. Clearly, we can notice a significant improvement in quality when incorporating a textual prior in feature inversion.


\end{document}